\RequirePackage{fix-cm}
\documentclass[twocolumn]{svjour3}          % twocolumn
\smartqed  % flush right qed marks, e.g. at end of proof
\usepackage{graphicx}
\usepackage{apalike}
\usepackage{setspace}
\usepackage{booktabs} % For formal tables
\usepackage{pdfpages}
\graphicspath{{images/}}
\usepackage{color}
\usepackage{wrapfig}
\usepackage{hyperref}
\usepackage{epstopdf}
\usepackage{subcaption}
\usepackage{array}
\usepackage{balance}
\usepackage{gensymb}
\usepackage{mathpazo}
\usepackage{amsmath}
\begin{document}

\title{Augmenting Visual SLAM with Wi-Fi Sensing For Indoor Applications%\thanks{Grants or other notes
}


\author{Zakieh S. Hashemifar$^{*}$         \and
        Charuvahan Adhivarahan$^{*}$       \and 
        Anand Balakrishnan$^{\dagger}$           \and
        Karthik Dantu$^{*}$  
}


\institute{$^{*}$ \at 
              Computer Science and Engineering Department, University at Buffalo, 338 Davis Hall, Buffalo, NY 14222. \\
              Fax: +1716-645-3484\\
              \email{\{zakiehsa, charuvah, kdantu\}@buffalo.edu}           %  \\
           \and
           $^{\dagger}$ \at
              Computer Science Department, University of Southern California, 941 Bloom Walk, Los Angeles, CA 90089. \\
              \email{anandbal@usc.edu}
}

\date{Received: date / Accepted: date}

\maketitle

%\vspace{-102pt}
\begin{abstract}
Recent trends have accelerated the development of spatial applications on mobile devices and robots. 
These include navigation, augmented reality, human-robot interaction, and others. 
A key enabling technology for such applications is the understanding of the device's location and the map of the surrounding environment. 
This generic problem, referred to as Simultaneous Localization and Mapping (SLAM), is an extensively researched topic in robotics. 
However, visual SLAM algorithms face several challenges including perceptual aliasing and high computational cost. 
These challenges affect the accuracy, efficiency, and viability of visual SLAM algorithms, especially for long-term SLAM, and their use in resource-constrained mobile devices. \\
A parallel trend is the ubiquity of Wi-Fi routers for quick Internet access in most urban environments.
Most robots and mobile devices are equipped with a Wi-Fi radio as well. 
We propose a method to utilize Wi-Fi received signal strength to alleviate the challenges faced by visual SLAM algorithms. 
To demonstrate the utility of this idea, this work makes the following contributions: (i) We propose a generic way to integrate Wi-Fi sensing into visual SLAM algorithms, (ii) We integrate such sensing into three well-known SLAM algorithms, (iii)  Using four distinct datasets, we demonstrate the performance of such augmentation in comparison to the original visual algorithms and (iv) We compare our work to Wi-Fi augmented FABMAP algorithm. Overall, we show that our approach can improve the accuracy of visual SLAM algorithms by 11\% on average and reduce computation time on average by 15\% to 25\%.
\end{abstract}

%\vspace{-5pt}
\section{Introduction}
\label{sec:intro}
%Rapid developments in computing, sensing and actuation have catalyzed the emergence of technology in our daily lives. 
%Particularly, there is a dramatic increase in the use of sensing to automate mundane tasks and/or enable novel applications to improve our lifestyle. 
%Due to miniaturization in computing and reduction in prices of cameras, a lot of such sensing is visual sensing. 
%Cameras are enabling robots to be autonomous and assist in various tasks such as vacuuming our floors, act as tour guides in museums, drive our cars and manage inventory at warehouses. 
%Similarly, cameras are being embedded in mobile devices such as smartphones, smart glasses and our gaming devices to enable augmented reality applications. 
%Most such applications require a constant perception of the environment and spatial reasoning of the robot/mobile device with respect to that environment. 
%%In robotics literature, these are referred to as the coupled problems of localization and mapping. 
%Simultaneous Localization and Mapping (SLAM)~\cite{thrun:probotics} is the joint estimation of both a representation of the environment and the robot's position (state) with respect to the environment.  
%
%\begin{figure}
%\centering
%\includegraphics[width=0.2\textwidth]{intro_images/rtabmap.png}
%\includegraphics[width=0.2\textwidth]{intro_images/orbslam.png}
%\caption{Two different representations of an environment. RGBD SLAM and RTAB-Map produce dense reconstructions (left) while ORB-SLAM generates sparse point clouds. (right)}
%\label{fig:intro_maps}
%\vspace{-19pt}
%\end{figure}
%need for SLAM
Recent technology trends have enabled the deployment of robots and mobile devices in urban areas for applications such as telecommuting, augmented reality, service robotics, and others. Most such applications require spatial reasoning - identifying the device location as well as recognizing parts of the surrounding environment. In robotics literature, these are referred to as the coupled problems of localization and mapping - jointly called Simultaneous Localization and Mapping (SLAM). SLAM has been extensively researched in the last two decades. 
Recent trends in sensing have seen the use of regular and depth cameras together (with sensors such as Microsoft Kinect) for 3D mapping. 
RGBD SLAM~\cite{burgard:rgbd-slam}, RTAB-Map~\cite{rtabmap} and ORB-SLAM~\cite{orbslam} are more recent examples. 
Of particular interest to this work is visual SLAM in indoor environments.
Algorithms reasoning with RGBD sensors come with some challenges when performing SLAM indoors. \\
%\kar{for each of these problems, we should cite some papers that also say these are problems}. \\
Some of the common problems are:\\
\textbf{Perceptual Aliasing}~\cite{perceptual_aliasing1,visual_wifi_2}\textbf{:} Indoor environments tend to be symmetric and repetitive. Corridors with bland walls and repeated patterns of doors and lights could potentially cause confusion between different similar places (wrong loop closure) resulting in faulty maps and bad localization. \\
%Further, if the robot/mobile device loses position information (kidnapped robot), it might be extremely challenging to re-localize. \\
\textbf{Computational Complexity}~\cite{computational_complexity1,memory_management2}\textbf{:} Cameras usually produce large volumes of data. 
For example, MS Kinect has a frame rate of 30 fps and each frame has more than 300000 points including color and depth data. 
This makes feature detection, matching and loop closure computationally more challenging, especially on resource constrained devices and over long runs.
%Extracting and matching features between depth/RGB images are computationally intensive tasks. 
%These comparisons need to be constantly made to estimate visual transformations and detect loops for correcting errors in the constructed map. 
%In a graph-based SLAM algorithm (such as RGBD SLAM), the number of images for potential comparisons increase with time making it harder to run them online, especially over long runs. 
%This is cause for innovation of several novel SLAM algorithms (such as RTAB-Map~\cite{rtabmap}). 
%Wi-Fi Access Points are ubiquitous in most urban environments including airports, offices, malls, and homes. Most robots are equipped with Wi-Fi cards for connectivity. There has been some prior work on using Wi-Fi sensing for localization, and the use of this coarse localization to assist visual SLAM. However, there is no generic way to demonstrate the utility of Wi-Fi sensing to assist visual SLAM, which is the goal of this work.
%As described above, a challenge for most visual SLAM algorithms is perceptual aliasing. 
%Particularly in repetitive environments, relying completely on visual information for place recognition can cause faulty loop closures leading to poor maps and localization. 

In a parallel trend, Wi-Fi radio is available on most robots or mobile devices and Wi-Fi Access Points are ubiquitous in most urban environments.
Wi-Fi and visual sensing are complementary to each other. While Wi-Fi sensing is less reliable than visual sensing, it is immune to perceptual aliasing.
The degree of detail in Wi-Fi data is much less than visual data and therefore requires much less processing time. 
%These approaches are not utilized for saving computation time.
%We hypothesize that Wi-Fi sensing is complementary to visual sensing, and resultant mapping/localization is correspondingly more accurate. 
%Further, we could alleviate the computational complexity of loop closures by restricting such feature matching using Wi-Fi sensing resulting in improved computation times of SLAM algorithms. 
In this work, we present a generic workflow to incorporate Wi-Fi sensing into visual SLAM algorithms in order to alleviate perceptual aliasing and high computational complexity. The contributions of this work are as follows:
%\vspace{-8pt}
\begin{itemize}
\item We propose a general way to integrate Wi-Fi sensing with visual SLAM by using received signal strength as an indicator of coarse spatial locality. Unlike many other methods, our integration works in tandem with the visual SLAM operation without any requirement of prior Wi-Fi data collection phase. 
\item We instrument three separate open-source visual SLAM systems (RGBD-SLAM, RTAB-Map, and ORB-SLAM) using our proposed technique to show the generality of our method.
\item We run our algorithm on four datasets from four different buildings to experimentally demonstrate the benefits of augmenting the three SLAM systems with Wi-Fi sensing on these four datasets.
\item We compare our work with the most recent state-of-the-art in WiFi-augmented visual sensing work which is Wi-Fi augmented FABMAP~\cite{visual_wifi_2}.
\end{itemize}

%The rest of the paper is as follows. In Section~\ref{sec:relwork}, we discuss prior work in the use of Wi-Fi for localization and mapping. 
%\ref{sec:wifi} introduces how we use Wi-Fi as a sensing modality. Our metric for Wi-Fi similarity, and ways we integrate it in the three visual SLAM algorithms is detailed in Sections~\ref{sec:slam} and ~\ref{sec:soln}. Section~\ref{sec:eval} presents the benefits of augmenting visual SLAM with Wi-Fi using the five datasets we collected. 
%Then, in \ref{sec:discussion}, we discuss implications of augmenting visual sensing with Wi-Fi. Finally, we conclude with thoughts on future work in Section~\ref{sec:conc}. 

%\vspace{\vertspcpostsec}

%\vspace{-5pt}
\section{Related work}
\label{sec:relwork}
%\vspace{\vertspcposthead}
%\zaki{In this section, we review existing work in literature which relate to our contributions.} 
There has been a lot of research on SLAM. We will present representative work. 

\textbf{Visual SLAM:} Prior work on SLAM has been done with multiple sensors including RGB and RGBD cameras, 2D and 3D LiDARs, 2D and 3D sonar sensors~\cite{burgard:rgbd-slam,cartographer}. 
%Prominent among these are visual SLAM systems that use visual features to match observations.% and further use these matches to estimate transformations between robot poses and landmarks. 
%SIFT~\cite{sift}, SURF~\cite{surf} and ORB~\cite{orb} are popular feature detectors used in visual SLAM algorithms. 
%While these visual features are generally designed to be as sparse as possible so as to manage execution time, their extraction and matching dominates the computation time in the end. 
%Also, all visual feature matching algorithms suffer from perceptual aliasing which causes failure in visually repetitive places. 
%This is common in some indoor environments. 
%SLAM using RGBD cameras has become prevalent in recent years and several SLAM algorithms are designed to use such sensors. 
A recent trend has been the use of color images with depth images. Some of more well-known visual SLAM examples include RGBD SLAM~\cite{burgard:rgbd-slam}, RTAB-Map~\cite{rtabmap}, and ORB-SLAM~\cite{orbslam}. 
Since we instrument these algorithms, they will be discussed in detail in Section~\ref{sec:soln}.
%These are graph-based SLAM algorithms with different approaches toward loop closure detection.~\cite{burgard:rgbd-slam} performs a heuristic search for frame selection for comparison,~\cite{rtabmap} keeps a list of frames which are believed to be useful for loop detection based on their proximity to current frame or frequent number of common features, and~\cite{orbslam} creates an inverted index of visual words to frames for retrieval of all frames having common words with current frame. 

\textbf{Wi-Fi Localization/SLAM:} 
%Capitalizing on the ubiquity of the 802.11 wireless technology, some localization and mapping methods using COTS wireless cards as sensors have been proposed. 
In robotics literature, there has been research on Wi-Fi localization~\cite{biswas-icra10,loc-unc-5} as well as Wi-Fi SLAM~\cite{loc-unc-7,thrun:wifi-slam,wifi-n-slam-2}. 
%In~\cite{thrun:wifi-slam}, the authors propose a method to perform SLAM using only Wi-Fi sensors and achieve a localization accuracy of 1.75m to 2.18m.%
%~\cite{biswas-icra10} models the world as a WiFi signal strength map with geometric constraints and uses Wi-Fi sensing as a continuous sensor to update robot location. 
%~\cite{loc-unc-6, wifi-n-slam-2} propose methods to perform localization based on Wi-Fi signal strength using Wi-Fi fingerprinting and Gaussian Processes. 
Similarly, wireless localization is a hot topic in mobile computing.~\cite{yang-mobicom12,luo-ipsn14} use wireless fingerprinting for localization. 
SpotFi~\cite{kotaru-sigcomm15}, Monoloco~\cite{soltanaghaei-mobisys18} and \cite{karanam_ipsn18} use channel state information (CSI) to achieve decimeter-level localization. 
\cite{kumar2018indoor} employ CSI and inertial measurements for delivering more accurate localization. 
%Another proposal~\cite{wifi-n-slam-1} adapts a Bayesian framework for SLAM using Inertial Measurement Units and Wi-Fi sensing. 
%In~\cite{signal-slam}, the authors propose using Wi-Fi along with other sensor data such as Bluetooth, magnetic field magnitude, NFC etc., to build signal maps for localization of people. 
%Similarly, there has been significant research on wireless localization in the mobile systems and mobile networking literature. 
%In the interest of space, we cite some representative ones.
%~\cite{chintalapudi-mobicom10} addressed the challenge of pre-deployment effort for wifi localization. 
%The authors propose a central EZ localization algorithm that incorporates the physics of wireless propagation, limited known positions and streams of wireless measurements from various devices in an indoor environment using a genetic algorithm to achieve accurate location information. 
%Zee~\cite{rai-mobicom12} improves this by using inertial sensors onboard mobile devices to reduce the calibration effort and crowd source WiFi fingerprints in a building. 

\textbf{Perceptual Aliasing:} %There has been significant effort to overcome perceptual aliasing in visual SLAM algorithms. 
\cite{cam-oscillations} incorporates a hardware solution to enhance feature measurements and localization by adding lateral motion to the camera.~\cite{sens-fus-2,sens-fus-3} combine information from multiple sensors for increasing localization and mapping accuracy.
%In~\cite{histogram-features}, RGB and gray-scale histograms are used as additional features for bounding visual keypoints comparison.
~\cite{feature_spatial_1} utilizes spatial uncertainties caused by actual measurements and image processing for better tracking.
Some approaches take into account different kinds of spatial information of image features in order to alleviate perceptual aliasing~\cite{feature_spatial_3}. %~\cite{feature_spatial_2, feature_spatial_3}. %caused by Bag Of Words methods
Recently, deep learning solutions are used for extracting better image descriptors which leads to more accurate place recognition~\cite{perceptual_aliasing1}.
%\cite{loc-acc-1} is in chinese languauge!
%\cite{loc-acc-2} didn't understand it that much!
%\cite{loc-acc-3} it uses some kind of artificial beacons on the sea surface for improved SLAM.
%\cite{feature_spatial_2} utilizes spatial information of image features for overcoming perceptual aliasing caused by BOW approaches.
%\cite{feature_spatial_3} uses relative spatial co-occurrence of words for overcoming perceptual aliasing caused by BOW approaches.
%\cite{fast-features} this one is related to visual tracking, not SLAM

\textbf{Computational Complexity:} Real-time performance is desirable in SLAM in some applications.
~\cite{memory_management2} and \cite{memory_management} incorporate memory management techniques to isolate a small active portion of the map to perform loop-closures quicker and ensure sustained online operation.
%\cite{memory_management} for handling large graphs in real-time.
~\cite{computational_complexity1} enables rapid multi-hypothesis testing in appearance-only SLAM using some probabilistic bail-out condition.

\textbf{Visual and Wi-Fi Integration:} Recent research has gravitated towards fusing alternate sensors, especially Wi-Fi, with visual sensing for improvements in localization accuracy.
In~\cite{loc-unc-5}, they model WiFi signal strength using a Gaussian process and use it for finding an initial seed estimate of the robot's location which is then refined with RGBD data.
~\cite{thrun:cam-wifi} utilizes a training phase for Wi-Fi modeling and then applies particle filters for fusing different sensors. % and \cite{stereo_RFID_1}
%Wi-Fi guided global localization~\cite{visual_wifi_1} collects Wi-Fi signatures and visual images of different unique places during an initial phase. Then, in a separate localization phase, Wi-Fi scan is compared to database entries and upon a successful Wi-Fi match, the corresponding images are compared.
%This work is the closest to our work, but they have not incorporated their work into any visual SLAM algorithm. 
%\cite{visual_wifi_2} incorporates Wi-Fi sensing in FABMAP~\cite{fabmap}. 
%They propose an approach for early fusion of Wi-Fi data and append the respective visual and Wi-Fi word vectors before running FABMAP. 
%In their approach, due to their agent's motion behavior and simplicity, no RSSI information is used and only the presence and absence of APs is investigated.
~\cite{visual_wifi_1,dong_sensys15} employ a mapping phase for collecting Wi-Fi signatures and visual images and utilize Wi-Fi data in localization phase for more accurate place recognition. 
~\cite{humanoid_wifi} uses Wi-Fi sensing along with other sensing modalities for more accurate localization estimates in less time. They incorporate sensing information directly in the map and use Monte Carlo estimator policy to allow point-matching only in nodes which is more probable to succeed.
While these research works take advantage of Wi-Fi data along with visual sensing for more accurate {\em localization}, none of them do SLAM and all of them employ an initial phase of Wi-Fi data collection or training which is later utilized for localization. 
%Further, they do not provide a general way to integrate wireless sensing with SLAM like our work does. This allows researchers to integrate wireless sensing into future SLAM algorithms as well as current ones. 

Closest to our work are~\cite{visual-wifi-ratslam,visual_wifi_3,visual_wifi_2}. % some research that integrate Wi-Fi and visual sensing
In~\cite{visual-wifi-ratslam}, authors incorporate Wi-Fi sensing only when visual data is no longer applicable. Our proposal is different in that it actively uses Wi-Fi data along with visual data and is generic making it applicable any visual SLAM algorithm.
%To demonstrate the improvements, we collect four datasets from four different buildings. We experimentally demonstrate the benefits of augmenting the three SLAM systems with Wi-Fi sensing on these four datasets. 
In \cite{visual_wifi_3}, the authors propose a voting based metric to fuse sensor information to find loop closures whereas our approach is different in that the visual matches are allowed only if Wi-Fi based matches are made to improve runtime efficiency.
FABMAP is augmented in \cite{visual_wifi_2} by tagging images with Wi-Fi vectors indicating the presence of APs. Apart from FABMAP being a topological SLAM, we use vectors of RSSI values instead of binary vectors. In section \ref{sec:eval}, we compare our approach with Wi-Fi augmented FABMAP in more detail.\\
Overall, we'd also like to note that none of these works provide a general way to integrate Wi-Fi sensing with SLAM as our work does. This allows researchers to integrate wireless sensing into future SLAM algorithms as well as current ones. 
% \kar{similar one sentence description of the next two papers also followed by a one line description of difference between our work and theirs. Lose the rest of the text. }
% In terms of contribution, \cite{visual_wifi_3} and \cite{visual_wifi_2} seem to be the closest to the method proposed in this paper. However, there are notable differences as we discuss in sections \ref{sec:eval} and \ref{sec:discussion}. Of these, we compare the performance of our method to the more recent state of the art Wi-Fi augmented FabMap~\cite{visual_wifi_2}. 

%While prior work has mostly used Wi-Fi sensing for improving localization and mapping in some specific ways, 
%Unlike prior work, our proposed method generalizes the potential use of wireless sensing, and deeply integrates such sensing with three visual SLAM systems. 
%This is done to demonstrate the general utility of Wi-Fi signal strength sensing, and how it could complement visual sensing for spatial applications. 
%In particular, we show that some of the challenges faced by visual SLAM systems such as localization/mapping accuracy and computational complexity could be alleviated with such integration. 
%In this paper, we propose a method to incorporate the signal information from wireless cards more directly as a pluggable part of RGBD SLAM algorithm to both construct a more accurate map and to reduce the execution time of the algorithm. Our method is complementary to other optimizations made to the RGBD SLAM algorithms.

%\vspace{\vertspcpostsec}

\vspace{-15pt}
\section{Wi-Fi sensing}
\label{sec:wifi}
Wi-Fi has become common in our lives and has enabled the mobility of our computing. Wireless sensing is a popular topic of research in mobile and sensor systems communities as well. Typical sensing includes received signal strength (RSSI). 
%Part of this movement is the deployment of Wi-Fi access points in most urban settings such as homes, offices, airports, malls, coffee shops and others. A popular trend is the use of Wi-Fi for sensing in addition to communication.
%Wi-Fi uses radio waves, typically in the 2.4 GHz or 5 GHz for communication. Several propagation models are proposed in literature that relate distance to the Received Signal Strength Indicator (RSSI) of the signal at the receiver. We wish to exploit this loose correlation in an indoor environment and augment our visual sensing for improved mapping.
Some modern Wi-Fi cards have the capability to calculate the wireless channel properties such as  Channel State Information (CSI). We chose to use Received Signal Strength (RSS) rather than CSI due to the following reasons: (a) CSI values were not available with APs at all locations that we measured, (b) Our preliminary CSI measurements did not demonstrate the accuracy reported by works such as SpotFI~\cite{kotaru-sigcomm15}, and (c) CSI empirically seems much more sensitive to small dynamics in the environment which affects robustness in our methodology.  
%We will now describe how we intend to use Wi-Fi RSSI with visual sensing. 
%and (d) CSI would provide too much data to efficiently process in the context of our work.
%\vspace{-5pt}
\subsection{Wi-Fi Similarity using Received Signal Strength}
%\vspace{-10pt}
As per the IEEE 802.11 standard, all Access Points (APs) constantly broadcast beacon frames that advertise their existence for potential clients. 
Additionally, all clients calculate the Received Signal Strength Indicator (RSSI) value for each AP that is visible to the client. 
RSSI is affected by many factors including distance, obstacles, and interference. Furthermore, each AP has an identifier called the Basic Service Set Identifier (BSSID), a value that must be unique to it for the functioning of any Wi-Fi network. %, which it broadcasts along with the beacon frames. 
BSSIDs can be read along with the RSSI values. % and be used as a convenient way of uniquely identifying APs. 

In modern urban environments, it is not uncommon to see fifteen to thirty BSSIDs at any given indoor location. This large number is due to different APs catering to different populations and providing access control to the network. On the robot/mobile device, RSSI values from multiple APs could be collected and used to construct a vector of RSSI values to form a {\em Wi-Fi signature}. Henceforth, in this paper, we use the term {\em Wi-Fi signature} and {\em signature} interchangeably. Such a vector is typically different at different locations that are sufficiently apart due to the fact that APs are usually spread out in space to maximize efficiency and connectivity. %Therefore, we conjecture that such a {\em Wi-Fi signature can be used as a coarse indicator of spatial locality}. 
While such Wi-Fi signatures have been used previously either as a lone sensor or combined with other sensing modalities for localization and/or mapping, our intent is to use them in an online fashion to augment existing SLAM algorithms for improved localization/mapping accuracy. 

\begin{figure}
\centering		
\includegraphics[width=.3\textwidth]{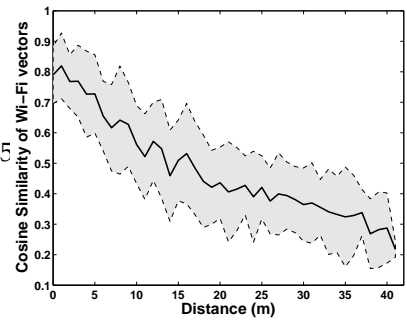}
\caption{Aggregate behavior of Wi-Fi Cosine Similarity against spatial distance as measured from various APs in B Hall. Results from other buildings were comparable and showed a similar trend}
\label{fig:wifi-similarity}
%\vspace{-17pt}
\end{figure}
%\vspace{-10pt}
To understand whether the similarity values are different at different locations, we collected Wi-Fi signatures at different points on a trajectory that starts to move away and return to the originating point. Specifically, we use Cosine Similarity between vectors as a measure of Wi-Fi similarity. 
Shown in Figure~\ref{fig:wifi-similarity} are aggregates of similarities in Wi-Fi vector space compared to the physical distance between locations from a building at our university. The trend clearly shows a coarse inverse correlation between physical distance and Wi-Fi similarity. This is expected since several models, including the ITU indoor propagation model, use this relation to find distance from RSSI values. 
%Similar observations regarding the spatial uniqueness of Wi-Fi signatures have been made by previous studies~\cite{biswas-icra10,thrun:cam-wifi} that perform indoor localization and we find them consistent with our analysis. %For example, \cite{yang-mobicom12} uses WiFi signatures to localize phones and \cite{biswas-icra10} uses them for both localization and navigation of mobile robots.
%\vspace{-5pt}
\subsection{Wi-Fi data processing}
{\bf Wi-Fi BSSID Dynamics: }An observation in RSSI aggregation is that each unique Access Point(AP) advertises different BSSIDs for different frequencies, 2.4GHz and 5GHz, and different access control mechanisms. These BSSIDs are only different in the last nibble of their MAC addresses. We aggregate all measurements from an AP by averaging the signal strength measured for all BSSIDs from a single AP.\\
{\bf Wi-Fi Data Collection:} The nature of Wi-Fi cards requires us to be stationary at a given location to collect steady signal strength readings. Therefore, during our measurements, our robot pauses for about ten seconds every few meters to collect Wi-Fi signals. \\
{\bf Wi-Fi Similarity Metric:}  The similarity between two different RSSI vectors $v$ and $w$ equals
%\begin{equation*}
$Similarity=\frac{v\cdot w}{\lvert v \rvert\lvert w \rvert}$. We use the cosine similarity measure in this work because it is invariant to scale. It is less sensitive to RSSI fluctuations due to configuration changes.
%\end{equation*} 

\section{General approach}
\label{sec:slam}
\begin{figure}
	\centering
	\includegraphics[width=0.3\textwidth]{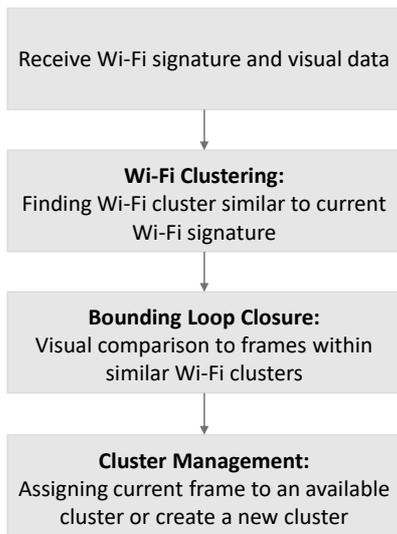}
	\caption{General approach to incorporate Wi-Fi into any visual SLAM algorithm.}
%\vspace{-20pt}
\label{fig:gen_approach}
\end{figure}

%\vspace{\vertspcposthead}
In Section~\ref{sec:wifi}, we described Wi-Fi similarity as a measure that correlates with coarse spatial locality. Our intent is to use this measure to improve visual SLAM. Figure~\ref{fig:gen_approach} shows the block diagram of our general proposed approach to incorporate Wi-Fi sensing into any visual SLAM algorithm. Following, we discuss each module in detail.

%\begin{itemize}
\textbf{Wi-Fi and Visual Association:} The first step is to associate a visual frame (image) with a corresponding Wi-Fi signature. For every 3 or 4 meters, the robot or mobile device pauses for about 10 seconds for Wi-Fi signature aggregation. Then, it associates any visual frame that follows with this Wi-Fi signature until the next pause.\\
\textbf{Wi-Fi Clustering:} Each Wi-Fi cluster represents a spatially separate region in the environment and has a representative Wi-Fi signature. 
It includes those frames which their signatures are similar to the representative signature and have at least one visual edge to another frame within the same cluster. In this module, we compute the cosine similarity between the Wi-Fi signature of the current frame and the representative signatures of all available Wi-Fi clusters within the database to find {similar clusters}. Any cluster within a threshold level of similarity is considered similar. These {similar clusters} represent spatial proximity to the current frame.\\
\textbf{Bounding Loop Closure Search:} A major challenge in SLAM is the problem of identifying a previously visited place. For example, if we go in a loop along the corridors of a building, we need to be able to recognize that we are back at the starting point once we complete the loop. This problem is called {\it Loop Closure}. As the map grows, SLAM algorithms accumulate many frames and it becomes computationally intensive to check for loop closures. Reducing the search space greatly benefits the timely working of a SLAM system. In this module, we reduce the search space by comparing the current frame only to frames within {similar clusters}. We do this to emulate comparison to frames from close-by regions.\\
\textbf{Cluster Management:} After permissible visual comparisons, the next step is to assign the current frame to the "correct" cluster. If any visual edge is added between the current frame and any frame within {similar clusters}, the current frame is assigned to that cluster. If there are multiple such clusters, the one with the highest cosine similarity is chosen. If no valid visual edges or {similar clusters} are found, a new Wi-Fi cluster is created and the Wi-Fi signature of the current frame is assigned as the representative signature of that cluster.

\section{Augmenting SLAM with Wi-Fi Sensing}
\label{sec:soln}
To demonstrate the utility of augmenting visual SLAM with Wi-Fi, we instrument three well-known SLAM algorithms with Wi-Fi sensing. For this, we chose RGBD SLAM~\cite{burgard:rgbd-slam}, RTAB-Map~\cite{rtabmap} and ORB-SLAM~\cite{orbslam}. All of them have open-source implementations making it convenient to modify them. In RGBD SLAM and ORB-SLAM, no odometry information is used which makes them easy to use on wearable devices as well as robots. We now describe each instrumentation in detail. For this, we first describe the original SLAM system followed by a description of our augmentation. 
\subsection{RGBD SLAM}
\subsubsection{\textbf{Background}}
RGBD SLAM~\cite{burgard:rgbd-slam} is a graph-based visual SLAM where nodes correspond to RGBD frames and edges correspond to 3D visual transformations between them. Also, any frame with unique visual features constitutes a keyframe.
RGBD SLAM represents an early SLAM system built for RGB-D sensors. It is somewhat brittle and computationally heavy as shown in our results.

\begin{figure}
%\vspace{-20pt}
\centering
\includegraphics[width=0.2\textwidth]{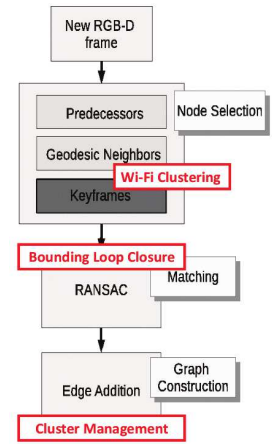}
\caption{Control Flow of RGBD SLAM for each new RGBD frame. Our augmentation of Wi-Fi sensing is shown in red}
\label{fig:RGBD_box}
\vspace{-15pt}
\end{figure}

Figure~\ref{fig:RGBD_box} shows a block diagram of RGBD SLAM. We also show how Wi-Fi sensing is augmented in red using the described modules in general proposed approach. 
In RGBD SLAM, each new frame is compared to a subset of previous frames for motion estimation and place recognition (Node Selection in Figure~\ref{fig:RGBD_box}). 
(a) {\it Predecessor Nodes}: A fixed number of nodes prior to the current node, (b) {\it Geodesic Neighbors}: A fixed depth of the graph before the current node, and (c) {\it Keyframes}: A randomized sub-set of previous keyframes. Intuitively, the predecessor nodes and geodesic nodes are used for motion estimation and the random sub-set of keyframes are for identifying long-term loop closures. Ideally, each frame should be compared with all relevant keyframes for best results. 
However, this is not tractable since the number of keyframes increases as the map grows. 
Thus, the keyframe selection is limited to a constant random number to reduce the computational complexity of this step as the map grows. This results in lower probability of finding correct long-term loop closures over time as the size of the graph increases.

\subsubsection{\textbf{Wi-Fi Augmentation}}
We intend to improve long-term loop closure (avoiding perceptual aliasing) by incorporating Wi-Fi sensing. As discussed earlier, %choosing a random set of keyframes results in poor loop closure detection and 
comparing all keyframes to the current frame results in huge computational overhead. It would be ideal to select a subgraph of the existing map that corresponds to the places close (distance-wise) to current frame for loop closure. This should improve the accuracy of loop closure detection along with runtime reduction. %while not increasing the computational complexity significantly. 

To improve the selection of related keyframes, we apply our proposed approach.
\begin{itemize}
\item \textbf{Wi-Fi Clustering:} We divide the RGBD keyframes into different clusters based on their Wi-Fi signature. 
For each new RGBD frame, we compute the cosine similarity between its signature and the representative signatures of all clusters to find {similar clusters}. 
\item \textbf{Bounding Loop Closure Search:} We compare the current frame only to RGBD keyframes within {similar clusters} for visual transformation calculation.
This approach limits the number of keyframes to be compared against for loop closure detection. 
\item \textbf{Cluster Management:} If the current frame is identified as an RGBD keyframe, this module is activated for assigning it to the right cluster as discussed in~\ref{sec:slam}.
\end{itemize}
Keyframe clustering based on the similarity between Wi-Fi signatures allows us to select a subset of RGBD keyframes that correspond to the similar location range as the current frame for loop closure detection. Also, due to low computation overhead of determining Wi-Fi similarity, we can compute similarities between the current Wi-Fi signature and {\it all} Wi-Fi clusters quickly, which is very beneficial in identifying keyframes from close-by regions from the whole map instead of picking random keyframes like the original method. This allows us to not only improve the loop closure accuracy but also decrease computational complexity. 

\subsection{RTAB-Map}
\subsubsection{\textbf{Background}}
RTAB-Map~\cite{rtabmap} is a real-time graph-based method similar to RGBD SLAM in structure, but it uses odometry. This makes it particularly suitable for robots or wearable devices with inertial sensing and low mobility. The main difference between RTAB-Map and RGBD SLAM is the way memory is managed and loop closure detection is handled.

\begin{figure}
\centering
\includegraphics[width=0.3\textwidth]{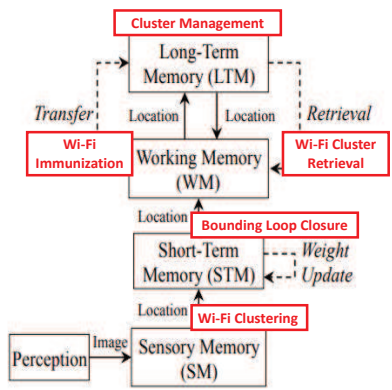}
    \caption{Block diagram of the workflow of RTAB-Map for each new RGBD frame~\cite{rtabmap} along with our Wi-Fi sensing augmentation in red.}
\label{fig:RTABMAP_flowchart}
\vspace{-20pt}
\end{figure}

In this algorithm, three types of memories are defined; Short Term Memory(STM), Working Memory(WM) and Long-Term Memory.
(a) STM is for a fixed number of the most recently added nodes %which would act as predecessor nodes for current node
, (b) WM includes the nodes which are candidates for comparison for loop closure detection. Every node is transferred from STM to WM after a while, and 
 (c) LTM is for the nodes which will not be used for any purpose anytime soon. 
 Figure~\ref{fig:RTABMAP_flowchart} shows the different memories and their interactions along with segments with our augmentation in red. We have built two additional modules called {\it Wi-Fi Immunization} and {\it Wi-Fi Cluster Retrieval} which are specific to this algorithm apart from those in our general approach. 

One of the parameters that a user can control in the RTAB-Map system is {\tt real\_time threshold}. This corresponds to the time that is considered acceptable for the processing of a new frame. During the execution, whenever the processing time of the current node exceeds the specified {\tt real\_time threshold}, some nodes are transferred from WM to LTM. Selection of nodes for transfer is done based on many criteria such as nodes that are not graph-wise close (i) to the current node and (ii) to nodes which have high visual similarity with the current node. While these conditions are reasonable, there is still a fair chance of losing relevant candidates, especially for long-term loop closure as discussed in the original paper~\cite{rtabmap}.
\subsubsection{\textbf{Wi-Fi Augmentation}}
RTAB-Map trades off localization and mapping accuracy for computational efficiency by moving frames from working memory (WM) to long-term memory (LTM). This process increases the probability of missing a loop closure due to unavailability of related frames in WM. We intend to improve the choice of frames to transfer to LTM using Wi-Fi sensing. 
\begin{itemize}
\item \textbf{Wi-Fi Clustering:} Upon arrival of a new frame, we compute the cosine similarity between the new Wi-Fi signature and all Wi-Fi clusters within memory in order to find {similar clusters}. 
\item \textbf{Wi-Fi Immunization:} If the RGBD frames of {similar clusters} are in WM, they are marked not to be moved to LTM. 
\item \textbf{Wi-Fi Cluster Retrieval:} If the RGBD frames of {similar clusters} are in LTM, they are retrieved back to WM and marked not to be moved to LTM.
\item \textbf{Bounding Loop Closure Search:} Visual transformation calculation happens between the current frame and the frames within {similar clusters}.
\item \textbf{Cluster Management:} The current frame is assigned to the right Wi-Fi cluster as discussed in~\ref{sec:slam}.
\end{itemize}  
\subsection{ORB-SLAM}
\subsubsection{\textbf{Background}}
\begin{figure}
%\vspace{-20pt}
\centering
\includegraphics[width=0.45\textwidth]{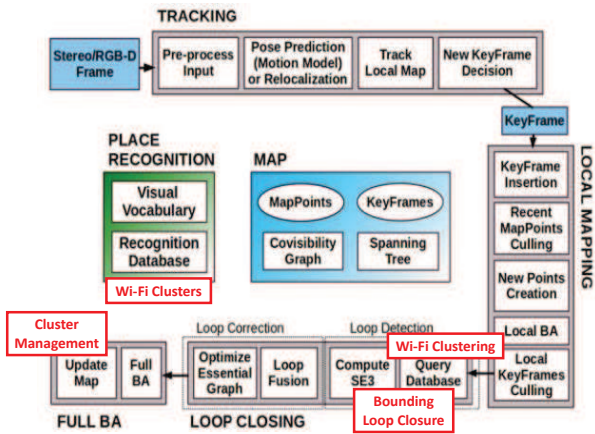}
    \caption{Control flow of ORB-SLAM for each new RGBD frame~\cite{orbslam} along with our incorporation of Wi-Fi sensing in red.}
\label{fig:orbslam_flowchart}
\vspace{-15pt}
\end{figure}

ORB-SLAM is a recent graph-based visual SLAM algorithm similar to RGBD SLAM in structure. A primary contribution of this algorithm is the construction of a dictionary relating visual words to keyframes which have observed them. This dictionary helps in quick lookup of similar keyframes (ones with similar visual words) for comparison to current keyframe for long-term loop closure. In this manner, each new keyframe is compared to the keyframes which have at least one visual word in common with it.
Another contribution is the definition of co-visible keyframes which identifies keyframes sharing map points. Figure~\ref{fig:orbslam_flowchart} shows the block diagram of ORB-SLAM along with segments where Wi-Fi modules are incorporated in red. 
Although the dictionary lookup approach increases the probability of accurate detection of positive loop closures, it still suffers from perceptual aliasing in symmetric environments.
\subsubsection{\textbf{Wi-Fi Augmentation}}
In ORB-SLAM, there is a visual word dictionary which maps visual words to RGBD keyframes which have observed them. For loop closure detection, each RGBD keyframe is compared to any keyframe which has at least one common visual word with it. While this approach probably would be able to find any correct loop closure, it is not immune to perceptual aliasing. Incorporating Wi-Fi sensing could alleviate this problem. Here is how we augment ORB SLAM with Wi-Fi sensing.
\begin{itemize}
\item \textbf{Wi-Fi Clustering:} We compute the cosine similarity between the Wi-Fi signature of new RGBD keyframe to all available Wi-Fi clusters to find {similar clusters}. 
\item \textbf{Bounding Loop Closure Search:} We only use the visual word dictionaries of {similar clusters} for finding RGBD keyframes which have visual words in common with the current frame. This would significantly lower the number of candidates for comparison and reduce the chances of perceptual aliasing.
\item \textbf{Cluster Management:} If a valid visual edge is constructed or there are any co-visible keyframes within {similar clusters}, the current frame is assigned to the corresponding cluster. Otherwise, a new Wi-Fi cluster with a separate visual word dictionary is created.
\end{itemize} 
We intend to open-source all implementations that we have augmented with Wi-Fi sensing on the publication of this work. We hope that this will help the community refine integrating Wi-Fi sensing and visual sensing. We will now evaluate the performance of each of our Wi-Fi augmented SLAM systems and compare them to the original approaches. 

%\vspace{-5pt}
\section{Evaluation}\label{sec:eval}
We now evaluate the computational complexity, localization and mapping accuracy of our Wi-Fi augmented SLAM algorithms. 
Section~\ref{subsec:slam-performance} describes the metrics used for evaluation in detail. 
In section~\ref{subsec:dataset}, we provide information about our setup for data collection and the collected datasets. 
Finally, we describe our results for all the datasets.
%\vspace{-5pt}
\subsection{Metrics to evaluate SLAM performance}
\label{subsec:slam-performance}
%\vspace{\vertspcposthead}
Here are some metrics that indicate SLAM performance: \\
\textbf{False Positive/Negative in Computing Loop Closure}: Identifying incorrect loop closures (false positive) and missing correct ones (false negative), especially in long-term, could have a huge effect on the accuracy of the constructed map. 
Based on ground truth knowledge of our datasets, we count the false positive and false negative loop closures for each SLAM algorithm. 
For this purpose, we find all loop closures that any well-constructed map has detected. Then any extra loop closure is counted as false positive and any missing one is counted as false negative.  \\
\textbf{Error in Estimated Trajectory}: 
%Since the estimated trajectory via SLAM is not necessarily in the same frame as ground truth, we use Kabsch algorithm~\cite{kabsch} for aligning the ground truth path and the estimated trajectory. 
As described below, we record ground truth for our datasets. To measure error, we use the Kabsch algorithm~\cite{kabsch} to align the trajectory generated
by the SLAM algorithm with the ground truth as they are not in the same coordinate frames. 
Then we calculate the RMS error between corresponding poses of ground truth trajectory and estimated trajectories
using the SLAM algorithm. \\
%Then, we calculate the error between the corresponding poses of ground truth trajectory and estimated trajectories using  
%\begin{equation} 
%    E = \frac{1}{N} \sum_{i = 1}^{N} \sqrt{(x_{i}^{(1)} - x_{i}^{(2)})^{2} + (y_{i}^{(1)} - y_{i}^{(2)})^{2}}
%\label{eq:error}
%\end{equation}
\textbf{Computation Time}: A principal challenge for SLAM approaches is the amount of time required to process the data.
 For all three algorithms, we have measured the difference in computation time between the original algorithm and our Wi-Fi augmented version. 
 More specifically, we micro-benchmark the difference in computation times for individual steps from our proposed approach. 
 This includes the reduction in computation time due to {\it Bounding Loop Closure} and the overhead resulting from {\it Wi-Fi Clustering} and {\it Cluster Management}.  
%\vspace{-5pt}
\subsection{Datasets}
\label{subsec:dataset}
Datasets with Wi-Fi measurements alongside RGBD measurements are not readily available. 
Hence, we collected four different datasets from four different buildings at our university.

For data collection, we used a Turtlebot\footnote{\url{https://www.turtlebot.com/turtlebot2/}} mounted with a Kinect 360 and a Velodyne VLP-16 LiDAR~\footnote{\url{http://velodynelidar.com/vlp-16.html}}. 
The Kinect 360 provides RGB-D data with RGB images of 640X480 resolution at 30 frames per second and depth images of 320X240 resolution at 30 frames per second. 
Its depth range goes from 1m to 4m approximately. The LiDAR provides 300,000 points per second with a 360\degree horizontal field of view and $\pm 15$\degree vertical field of view. 
It has a depth range of over 100m. 
In our datasets, the LiDAR is used for ground truth trajectory estimation. 
For this purpose, we used {\it Google Cartographer}~\cite{cartographer} for 2D ground truth trajectory estimation at a 5 cm resolution. %\kar{how much we are doing better? is it higher than this accuracy?}. 
All data is collected using ROS\footnote{\url{http://wiki.ros.org/}} and a laptop on the Turtlebot during execution. We use Intel 7265 wireless card for our Wi-Fi measurements.

We collected four separate datasets from four buildings on campus. \\
%\begin{itemize}
\textbf{C Hall:} This dataset was collected during traversal of a medium-sized square loop with corridors that are 20 meters long. There are 24000 images, less than 40 APs and a total of 7 Wi-Fi clusters in this dataset.\\
\textbf{B Hall:} This dataset includes one long and one short loop that links together to look like the number 8. There are 28000 images, about 40 APs and 13 Wi-Fi clusters in this dataset.\\
\textbf{J Hall:} This dataset includes one long loop and an adjoining trajectory which together looks like the number 9. There are a total of 19 Wi-Fi clusters, around 70 APs and 33000 images in this dataset.\\
\textbf{A Hall:} This dataset is one loop of a long jogging track with sparse visual features. There are not many blocking walls between different places on the trajectory. The number of frames, APs and Wi-Fi clusters are 50000, 45 and 8 respectively.
%\end{itemize}
%\subsubsection*{Statistics}
%\zaki{provide a table of number of clusters and frames of datasets in order to give some explanations about how number of clusters are affected.} 
%\vspace{-5pt}
\subsection{RGBD SLAM performance}
As we ran the RGBD SLAM with default parameter settings on {\it C Hall} dataset, we got very inaccurate maps and trajectories. 
So, we decided to perform parameter tuning and find the best possible outcomes of vanilla RGBD SLAM based on our metrics defined above. 
%The parameters are {\it min-matches} and {\it inlier-distance}.
The first parameter is the minimum number of matched features required for accepting a transformation which is called {\it min-matches}.
The second parameter is the maximum distance allowed for inlier points when using RANSAC for transformation estimation and is called {\it inlier-distance}.

%Increasing min-matches decreases the possibility of matches and has the effect of reducing false positives and increasing false negatives. 
%Increasing inlier-distance has the effect of increasing false positives and decreasing false negatives. 
We pick three different reasonable values for inlier-distance and four values for min-matches and run vanilla RGBD SLAM on each set of parameters for {\it C Hall Dataset}. 
\begin{figure*}
	\begin{subfigure}[b]{.3\textwidth}
		\includegraphics[width=\textwidth]{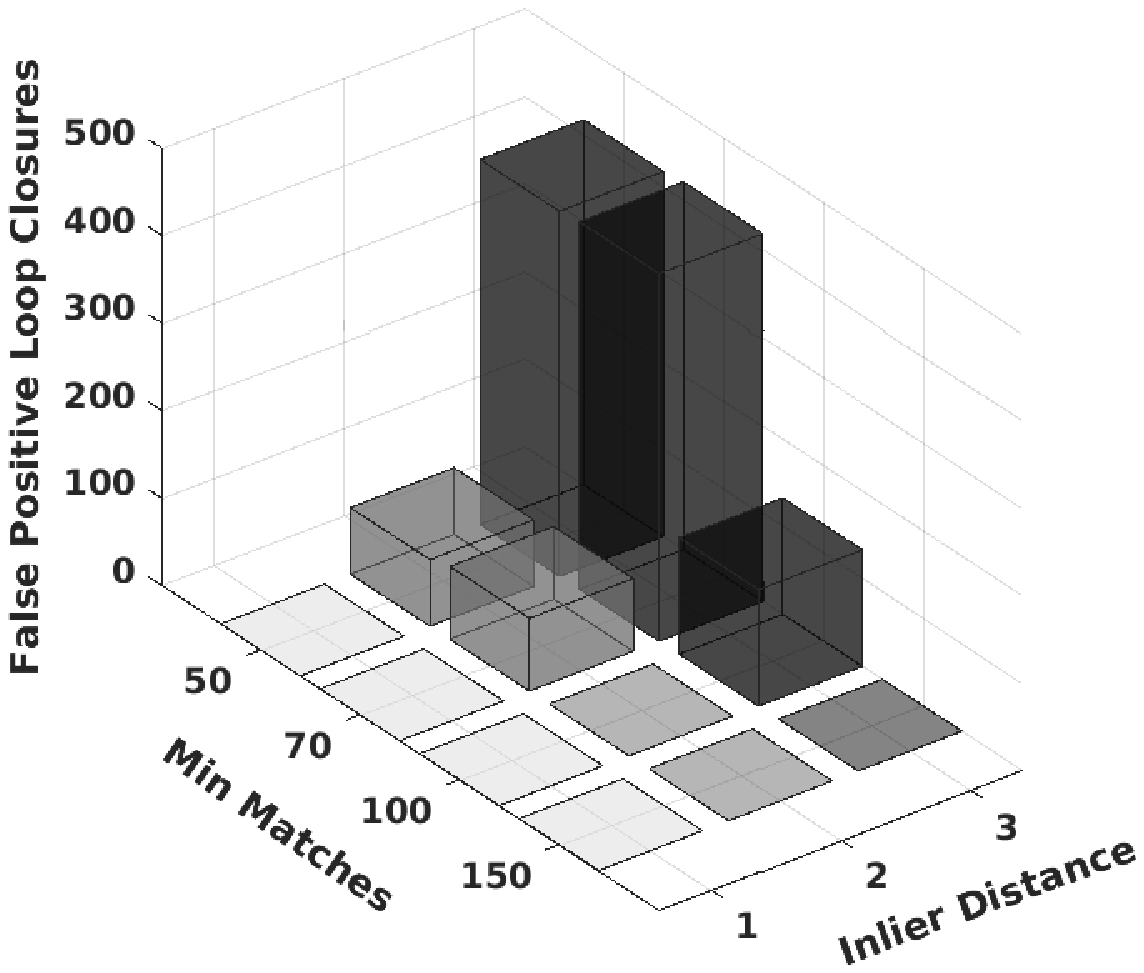}
		%\label{subfig:center}
		%\vspace{-6mm}
		%\caption{Number of false loop closures}
	\end{subfigure}
	\begin{subfigure}[b]{0.3\textwidth}
		\includegraphics[width=\textwidth]{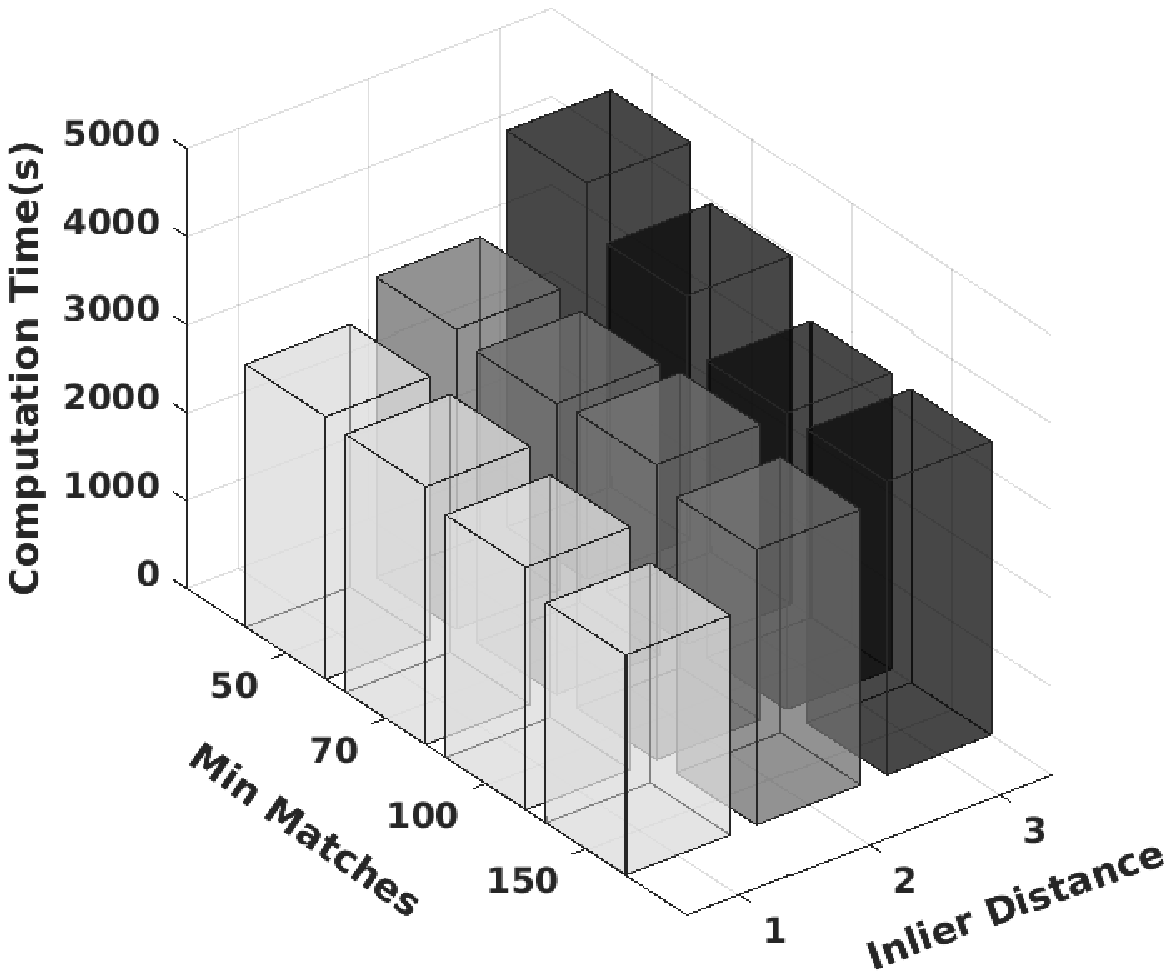}
		%\label{subfig:center}
		%\vspace{-6mm}
		%\caption{Number of keyframes}
	\end{subfigure}
	\begin{subfigure}[b]{0.3\textwidth}
		\includegraphics[width=\textwidth]{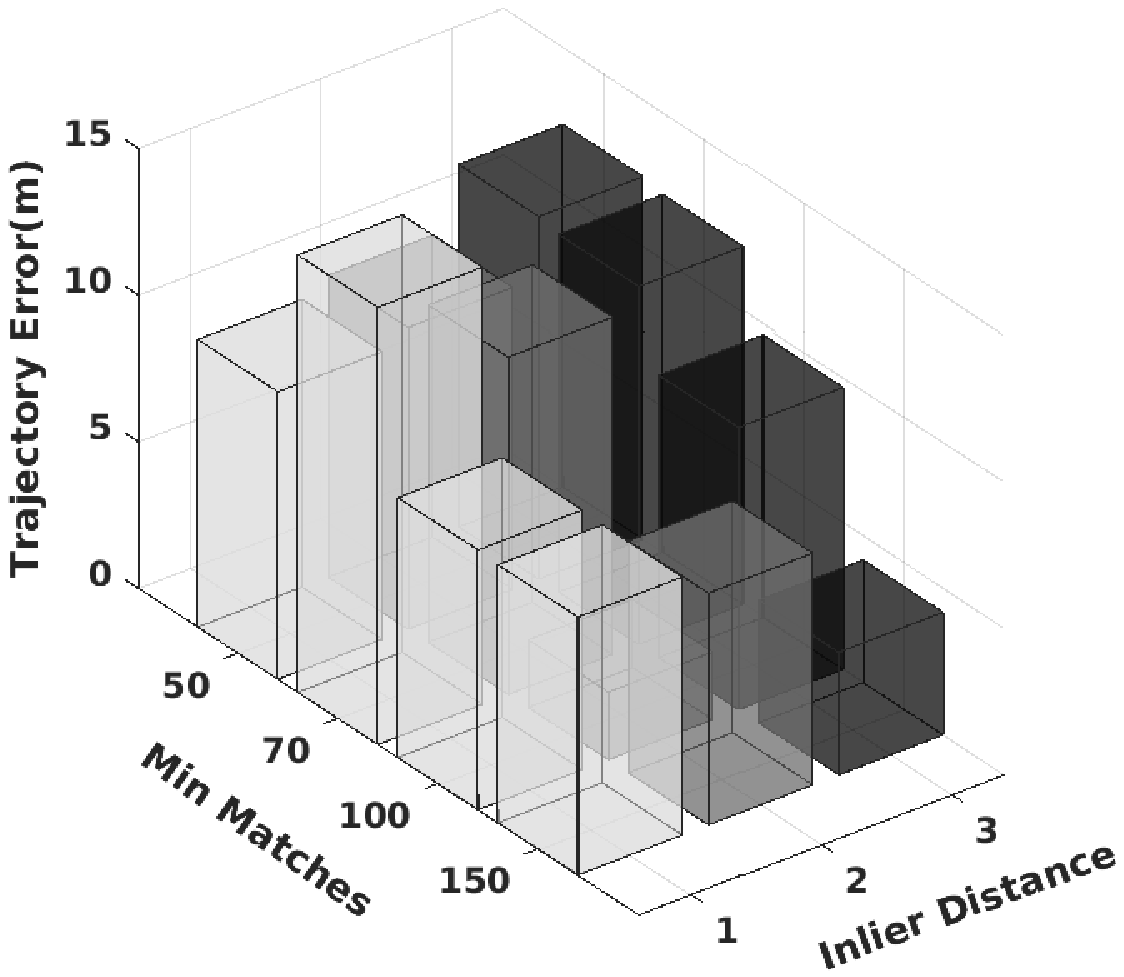}
		%\label{subfig:center}
		%\vspace{-6mm}
		%\caption{Estimated Trajectory Error}
	\end{subfigure}
\caption{False positive loop closures (left), computation time (center) and trajectory error (right) when the parameters {\it min-matches} and {\it inlier-distance} are varied. }
\label{RGBD_analysis}
%\vspace{-10pt}
\end{figure*}

%Figure~\ref{RGBD_analysis} shows the behavior of RGBD SLAM algorithm when two parameters are changing.
%One of these parameters is the minimum number of matched features required for accepting a transformation which is called \textit{min matches}.
%The other parameter is the maximum distance allowed for inlier points when using RANSAC for transformation estimation and is called \textit{inlier distance}.

%As it is expected, when \textit{min matches} increases, the number of keyframes increase as well.
%The reason is behind keyframe definition which states if a frame could not be related to any previous keyframe or any frame earlier than the last keyframe, it is a keyframe itself.
%On the other hand, increasing \textit{inlier distance} leads to decrement in the number of keyframes, because the probability of finding a transformation among two frames is higher.
%As observed in the leftmost plots, increasing \textit{min matches} and decreasing \textit{inlier distance}, usually decreases the number of false positive loop closures.
%Because increasing the \textit{min match} parameter makes it harder for the algorithm to find transformations having required number of matches among frames.
%In other words, it is less probable to find required number of matches within allowable inlier distance among two distant frames.

From Figure~\ref{RGBD_analysis}, increasing min-matches and decreasing inlier-distance decreases the number of false positives as expected. 
The computation time decreases as the number of min-matches increase and inlier-distance decreases. This is a direct result of the reduction in permissible transformations between frames.
There is a lack of structure in the way trajectory error varies with these parameters. 
Our conjecture is that this is because of the randomness in node selection as well as the effect of associated false positive/negative loop closures. 
%Note that the vanilla RGBD SLAM fails to perform the final loop closure (false negative) for all runs. 

\begin{figure}
\centering
\includegraphics[width=0.2\textwidth]{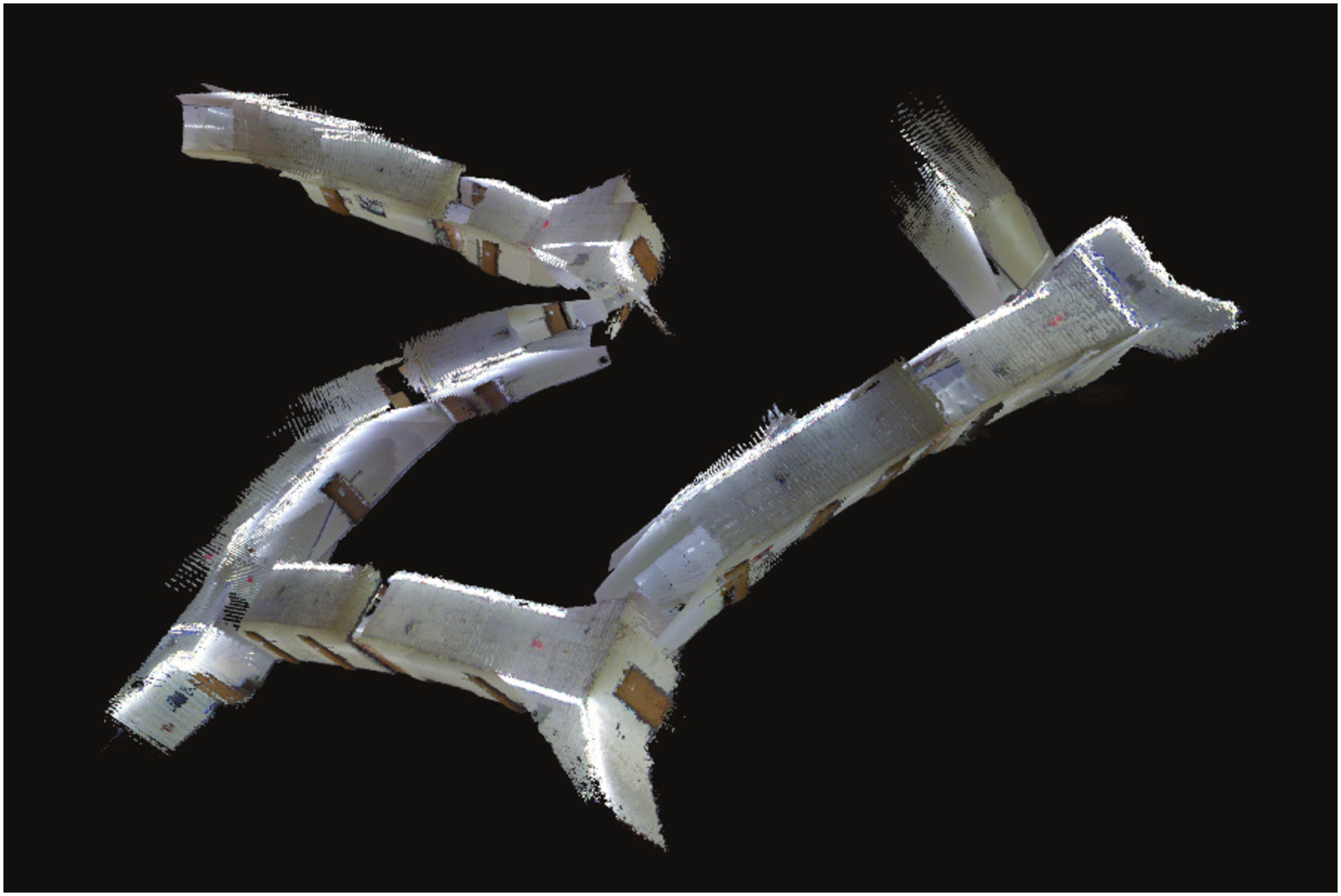}
\includegraphics[width=0.2\textwidth]{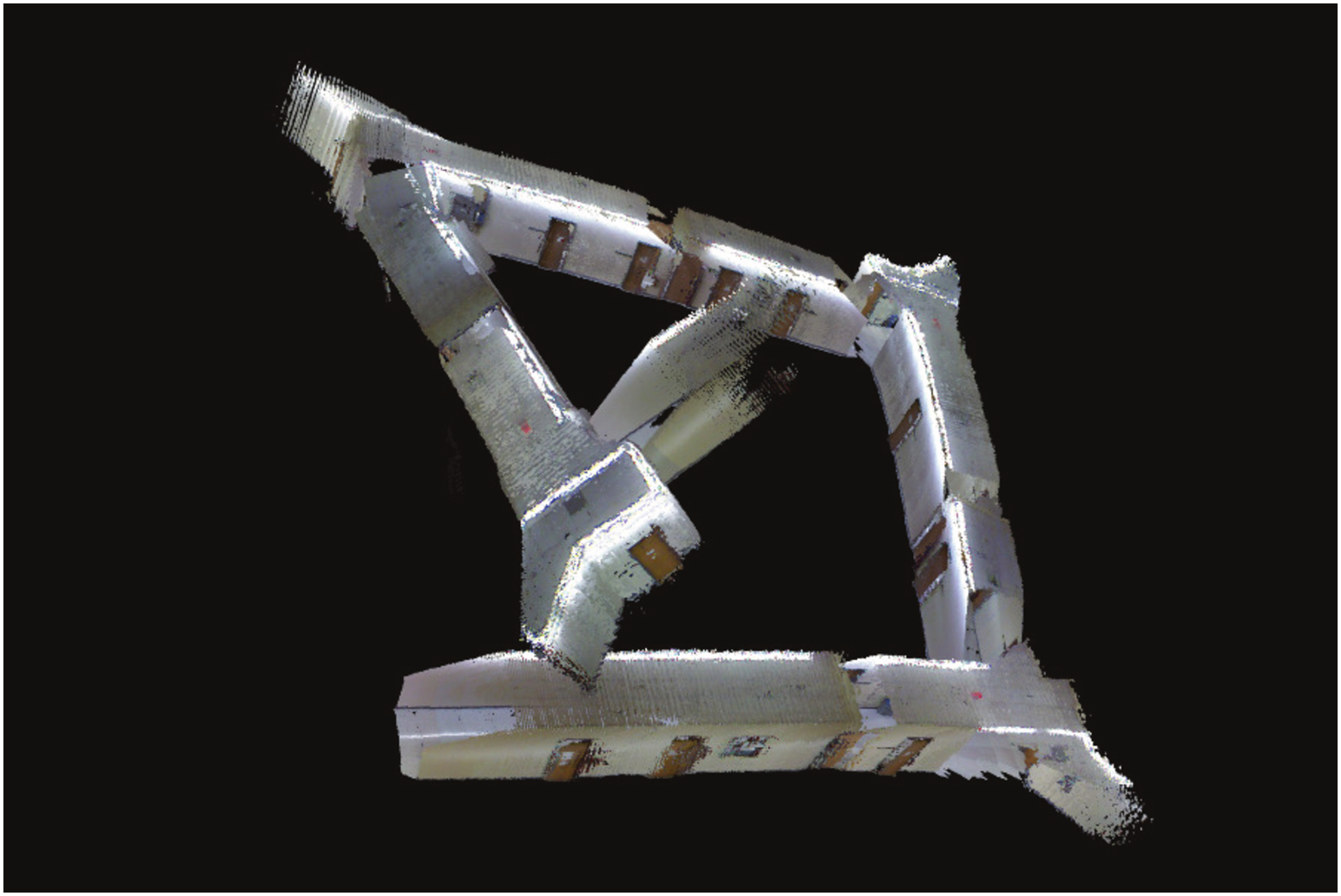}
\caption{Constructed maps of vanilla RGBD SLAM with {\it min-match}=100/150 and {\it inlier-distance}=2. Both suffer from rotational deviations due to low number of visual transformations}
\label{fig:rgbd_badmap}
\end{figure}
\vspace{-10pt}
Setting {min-match} to very high values and {inlier-distance} to very low values cause the graph to have a very low number of permissible visual transformations. 
Potentially too low to even create a reasonable map. 
Figure~\ref{fig:rgbd_badmap} shows the constructed maps with {min-match} = 100/150 and {inlier-distance}=2 which suffer from rotational deviations due to a very low number of visual transformations. 
Therefore, we have picked lower values for {min-match} and higher values for {inlier-distance} to provide parameters that give reasonable performance with respect to the number of visual transformations. 
Because Wi-Fi sensing is not able to force a visual transformation if it is rejected due to a very high value of {min-match} or very low value of {inlier-distance}.
%Figure~\ref{fig:RGBDW_maps} represents the two best constructed maps of our proposed method.
%As it is obvious, Wi-Fi incorporation is not only helping in getting rid of many false loop closures, but aids in detecting the final loop closure by making comparisons to related keyframes in close-by regions instead of non-sense random selection.
%It should be noted that the reason behind not very carefully aligned aisles at last loop closure is the way RGBD SLAM calculates visual transformations and doesn't relate to our method.
%However, it should be noted that for settings with very low number of false loop closures, the accuracy of estimated map is almost the same for both original and proposed approach.
%Because even our proposed method could not detect the last loop closure due to very low number of inliers in a feature-less environment.
%But our computation time is lower due to rejection of many transformations due to Wi-Fi dissimilarity among their corresponding frames prior to any effort for visual transformation.

\begin{figure*}
	\begin{subfigure}[b]{.3\textwidth}
		\includegraphics[width=\textwidth]{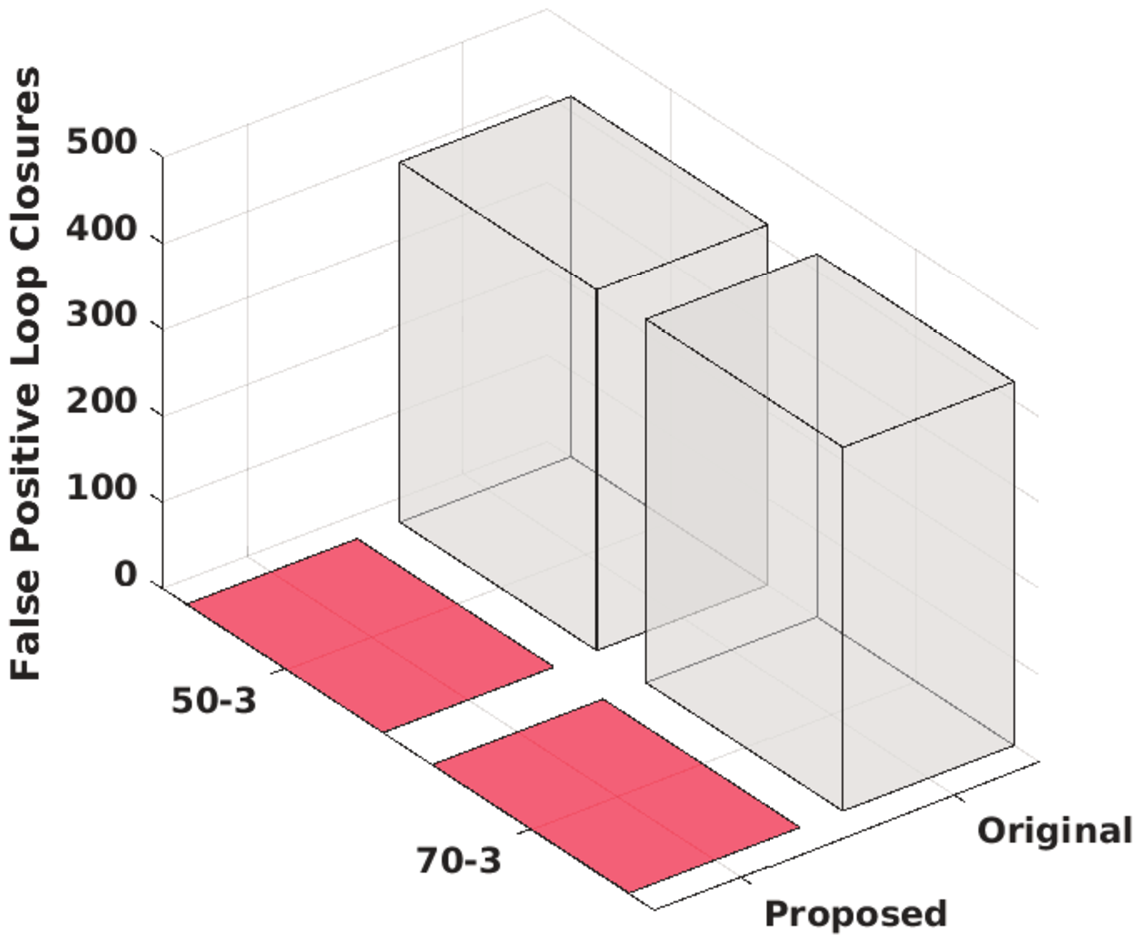}
		%\label{subfig:center}
		%\vspace{-6mm}
		%\caption{Number of false loop closures}
	\end{subfigure}
	\begin{subfigure}[b]{0.3\textwidth}
		\includegraphics[width=\textwidth]{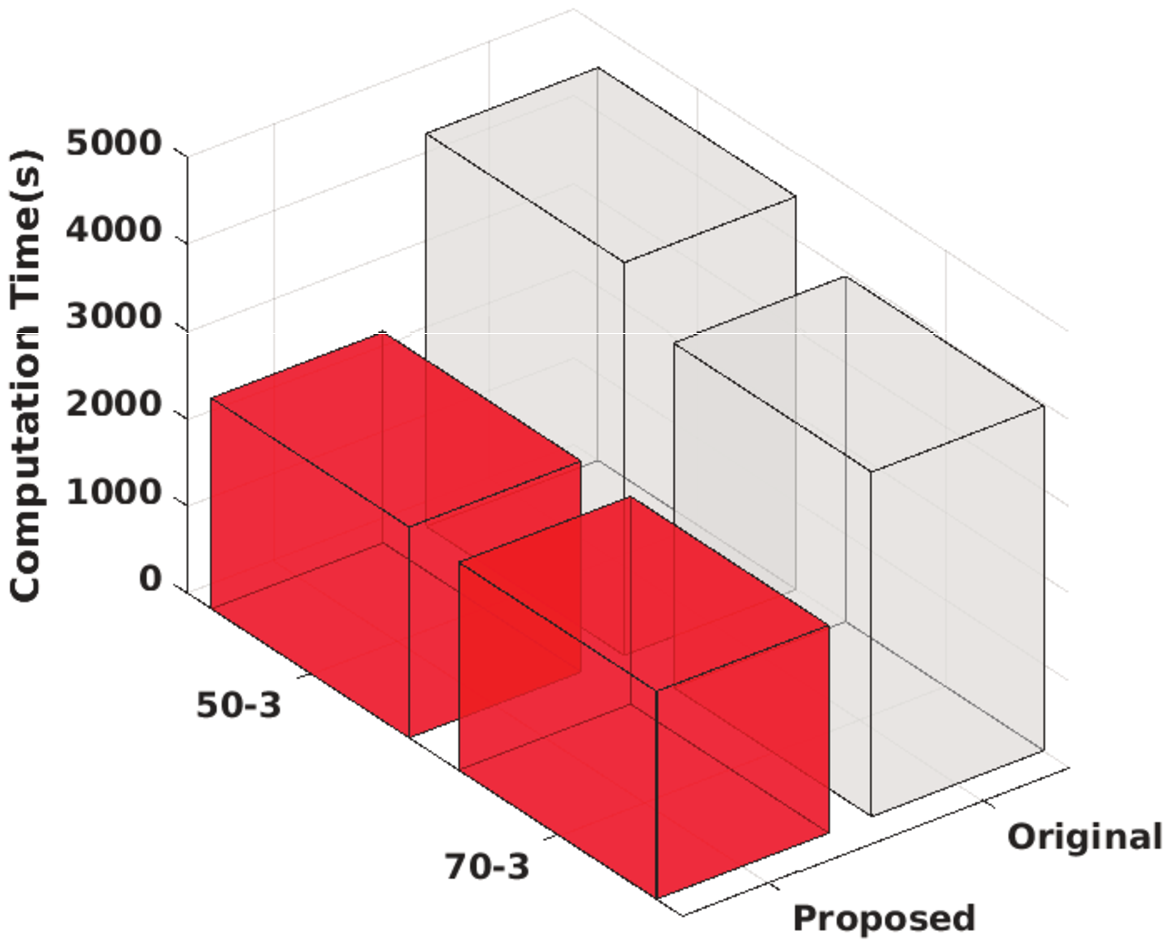}
		%\label{subfig:center}
		%\vspace{-6mm}
		%\caption{Estimated Trajectory Error}
	\end{subfigure}
%	\hspace{4mm}
	\begin{subfigure}[b]{0.3\textwidth}
		\includegraphics[width=\textwidth]{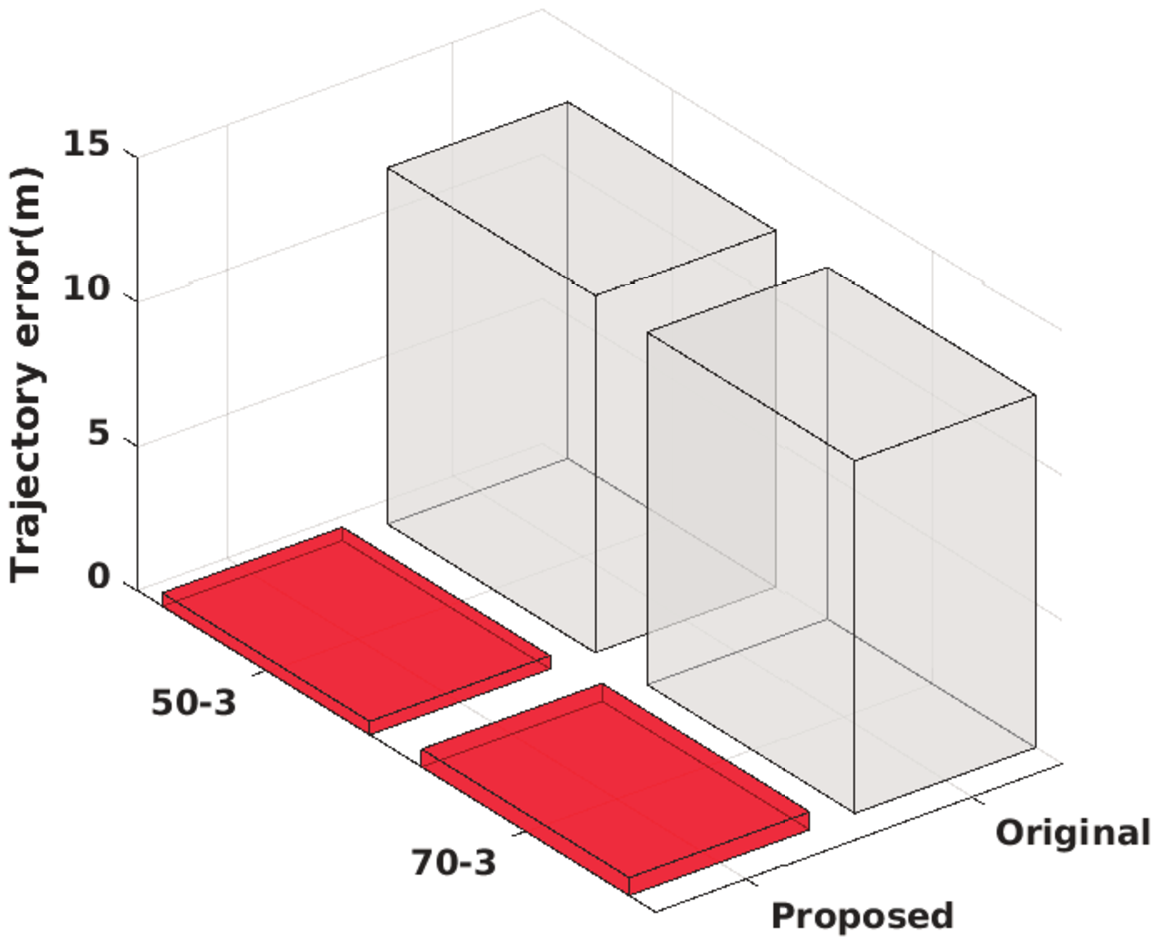}
		%\label{subfig:center}
		%\vspace{-6mm}
		%\caption{Number of keyframes}
	\end{subfigure}
%\vspace{-20pt}
\caption{Accuracy comparison between vanilla RGBD SLAM and our proposed method.}
\label{fig:RGBDW_comparison}
%\vspace{\vertspcpostimg}
%\vspace{-20pt}
\end{figure*}
Figure~\ref{fig:RGBDW_comparison} compares the original method and our proposed method for the C Hall dataset.
Results show that our proposed method gets rid of all false loop closures for both settings.%by more than $90\%$ for $min-matches=50$ and more than $80\%$ for $min-matches=70$.
We believe this is because Wi-Fi similarity allows us to identify a good sub-set of keyframes to compare against for accurate loop closure as opposed to the fixed set of random keyframes selected by vanilla RGBD SLAM. 
Similarly, we reduce the computation time of affected modules by {bounding loop closure} by more than $50\%$ for {min-match}=50 and more than $30\%$ for {min-match}=70. 
This is because Wi-Fi similarity allows us to eliminate unnecessary comparisons. 
Moreover, having a smoother map reduces the optimization time as well. 
Finally, we also observe a reduction in RMS error of $80\%$ for min-matches $\in (50, 70)$. 
This is also a result of identifying a good sub-set of keyframes for loop closure comparison and not allowing visual transformations between frames having distant Wi-Fi signatures. 

Based on the shown result, it seems that the performance of RGBD SLAM in symmetric environments with a low number of features is very poor. 
So we decided not to continue running this algorithm on longer and harder datasets in order to avoid unreasonable comparisons.  
%\vspace{-5pt}
\subsection{RTAB-Map performance}
%One main difference between RTABMap and RGBD SLAM is the real time constraint provided in RTABMap.

\begin{figure}[h]
    \centering
    \includegraphics[height=1.0in]{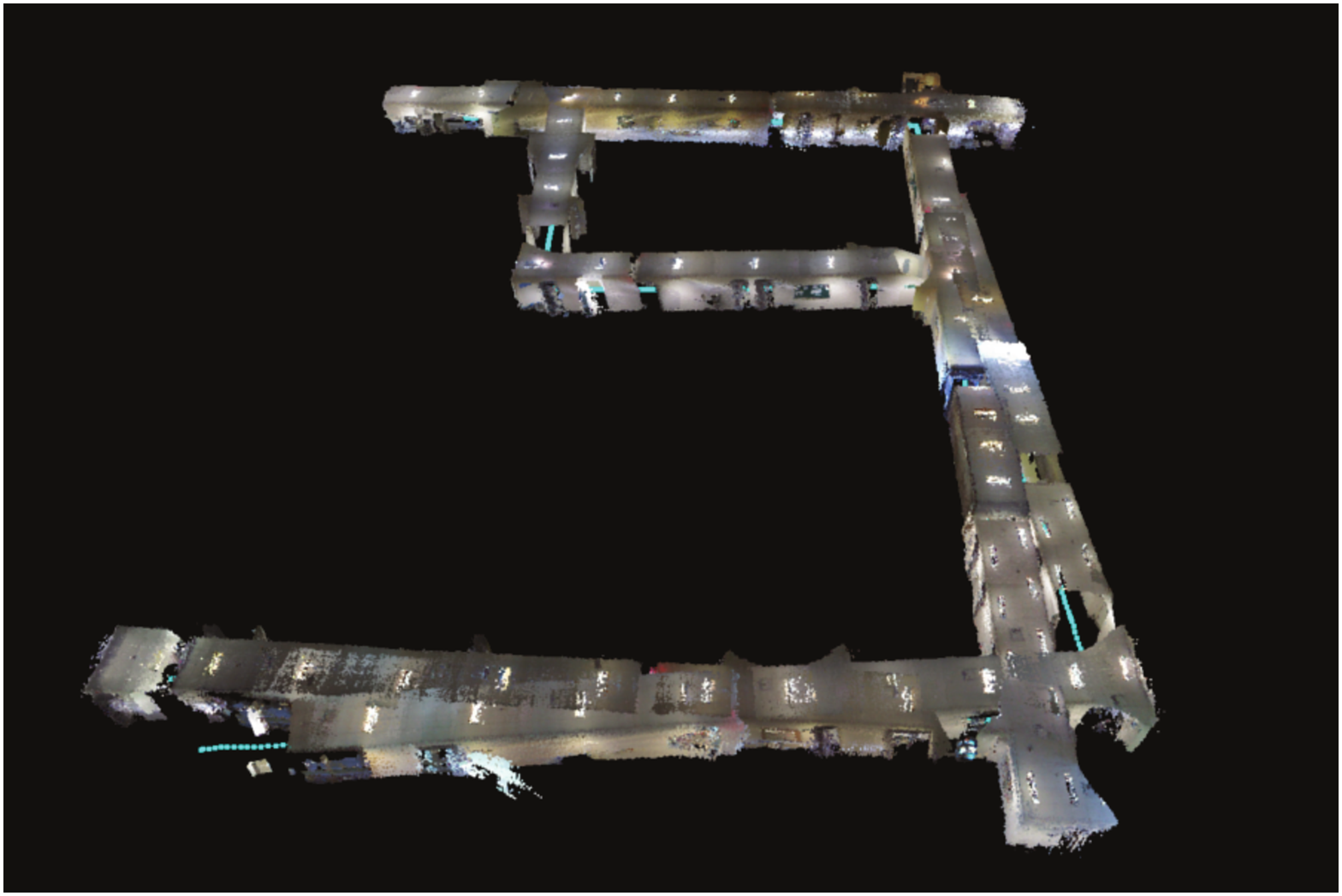}
    \hspace{.2in}
    \includegraphics[height=1.0in]{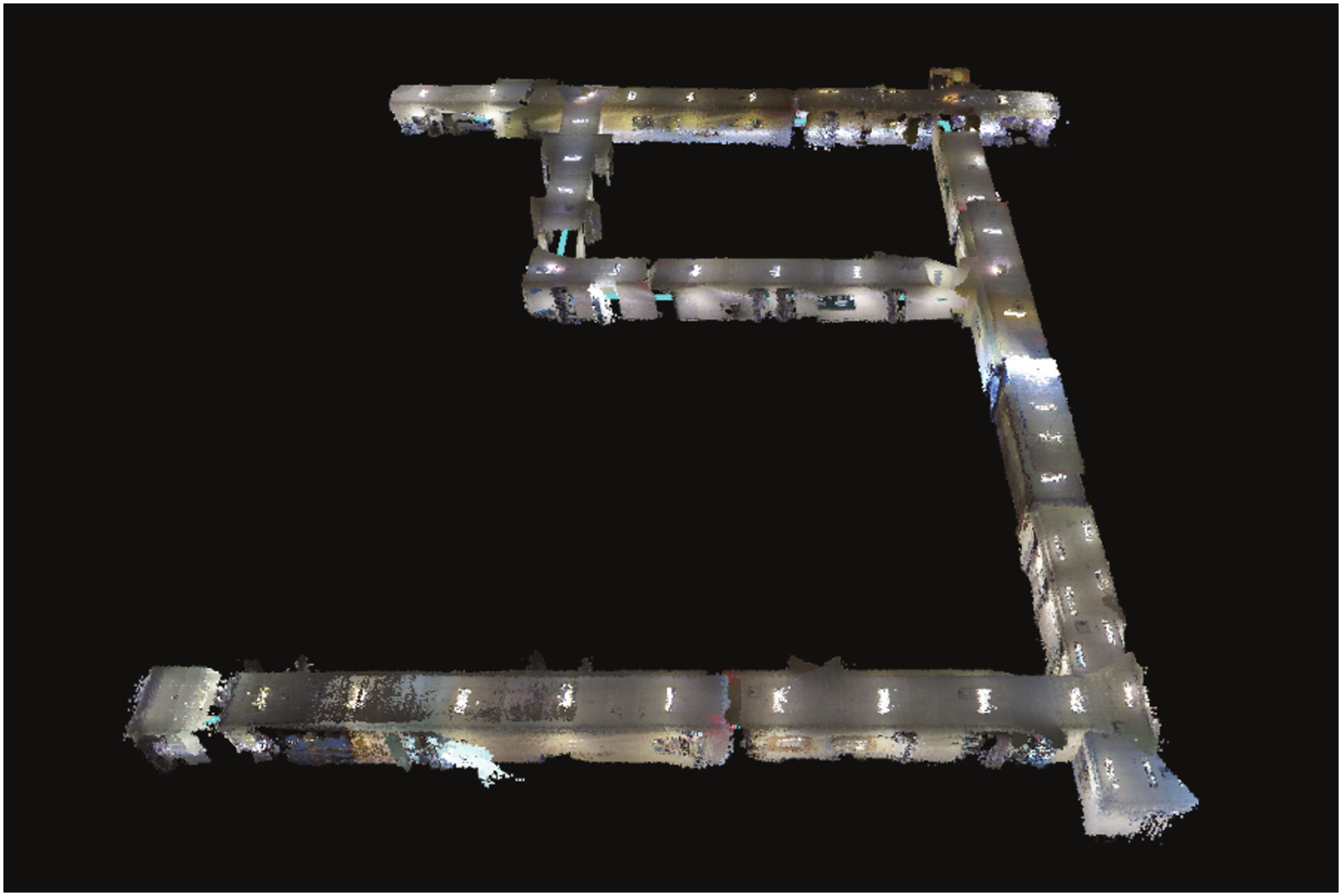}
    \hspace{.2in}
    %\includegraphics[height=1.0in]{Figure9_c.eps}
    %\hspace{.2in}
    %\includegraphics[height=1.0in]{Figure9_d.eps}
    %\hspace{.2in}
   % \includegraphics[height=1.0in]{Figure9_e.eps}
   % \hspace{.2in}
    %\includegraphics[height=1.0in]{Figure9_f.eps}
%\caption{Constructed maps of J Hall (top row) and A Hall (center row) using original RTAB-Map (left) and Wi-Fi augmented RTAB-Map (right).  The cropped images (bottom row) are the portion of the map from A Hall (center row) highlighted in the red rectangle. They show the difference in constructed map without loop closure in original approach (left) and with correct loop closures in Wi-Fi RTAB-Map (right)}
\caption{Constructed maps of J Hall (top row) using original RTAB-Map (left) and Wi-Fi augmented RTAB-Map (right). They show the difference in constructed map without loop closure in original approach (left) and with correct loop closures in Wi-Fi RTAB-Map (right)}
\label{fig:rtabmap_maps_2}
%\vspace{-10pt}
\end{figure}
\begin{table*}
\caption{{\bf RTAB-Map:} Trajectory error (m) for different datasets and different {\it real-time thresholds}}
\begin{center}
\begin{tabular}{| c | r r | r r | r r | r r | } 
\hline 
\multicolumn{9}{|c|}{Real-time Threshold (ms)} \\ 
\hline 
 & \multicolumn{2}{|c|}{$\infty$} & \multicolumn{2}{|c|}{70} & \multicolumn{2}{|c|}{100} & \multicolumn{2}{|c|}{200} \\ 
\hline 
 Dataset & \multicolumn{1}{|c|}{Vanilla} & \multicolumn{1}{|c|}{WiFi} & \multicolumn{1}{|c|}{Vanilla} & \multicolumn{1}{|c|}{WiFi} & \multicolumn{1}{|c|}{Vanilla} & \multicolumn{1}{|c|}{WiFi} & \multicolumn{1}{|c|}{Vanilla} & \multicolumn{1}{|c|}{WiFi} \\ 
\hline 
 A Hall & 0.2304 & 0.2284 & 0.4895 & 0.2537 & 0.4894 & 0.2846 & 0.2329 & 0.2074 \\ 
 B Hall & 0.2424 & 0.2262 & 0.1284 & 0.2219 & 0.1284 & 0.2219 & 0.1924 & 0.2200 \\ 
 C Hall & 0.1280 & 0.1100 & 0.1198 & 0.1005 & 0.1198 & 0.1005 & 0.1149 & 0.1416 \\ 
 J Hall & 0.3030 & 0.2113 & 1.3472 & 0.1945 & 1.3472 & 0.1999 & 0.2300 & 0.2132 \\ 
\hline 
\end{tabular} 
\label{table:rtabmap_rmse}
\end{center}
%\vspace{-10pt}
\end{table*}

\begin{table*}
\caption{{\bf RTAB-Map:} False negative loop closures for different datasets and different {\it real-time thresholds}}
\begin{center}
\begin{tabular}{| c | r r | r r | r r | r r | } 
\hline 
\multicolumn{9}{|c|}{Real-time Threshold (ms)} \\ 
\hline 
 & \multicolumn{2}{|c|}{$\infty$} & \multicolumn{2}{|c|}{70} & \multicolumn{2}{|c|}{100} & \multicolumn{2}{|c|}{200} \\ 
\hline 
 Dataset & \multicolumn{1}{|c|}{Vanilla} & \multicolumn{1}{|c|}{WiFi} & \multicolumn{1}{|c|}{Vanilla} & \multicolumn{1}{|c|}{WiFi} & \multicolumn{1}{|c|}{Vanilla} & \multicolumn{1}{|c|}{WiFi} & \multicolumn{1}{|c|}{Vanilla} & \multicolumn{1}{|c|}{WiFi} \\ 
\hline 
 A Hall & 0.0000 & 4.4444 & 100.0000 & 17.7778 & 100.0000 & 31.1111 & 11.1111 & 13.3333 \\ 
 B Hall & 8.0000 & 16.0000 & 100.0000 & 0.0000 & 100.0000 & 0.0000 & 6.0000 & 12.0000 \\ 
 C Hall & 6.0606 & 0.0000 & 100.0000 & 9.0909 & 100.0000 & 9.0909 & 27.2727 & 3.0303 \\ 
 J Hall & 9.0909 & 9.0909 & 100.0000 & 13.6364 & 100.0000 & 40.9091 & 0.0000 & 9.0909 \\ 
\hline 
\end{tabular} 
\label{table:rtabmap_false_negative}
\end{center}
%\vspace{-20pt}
\end{table*}
RTAB-Map is another state-of-the-art SLAM system. A key parameter in RTAB-Map is the ability to control the running time of the algorithm by setting {real-time threshold} parameter. 
The algorithm tries to keep the processing time of each node under {real-time threshold} by moving unused frames to LTM only to retrieving them back when necessary. 
In our experiments, we set different real-time threshold values and compare the behavior of vanilla RTAB-Map with our proposed approach for different datasets. 
We note that {real-time threshold}=$\infty$ means no threshold is set and no frame is moved to LTM.

Table~\ref{table:rtabmap_rmse} represents the trajectory error of both original and Wi-Fi augmented RTAB-Map for different datasets and {real-time thresholds}. 
For {real-time thresholds} of $\infty$ and 200, the trajectory error of both approaches is on the same order.
This is because of the number of frames moved to LTM. For a value of 0, no frame is moved to LTM. 
For a value of 200, the number of transferred frames is too low, because the processing time of most of the frames is less than this threshold.
Some of the error differences for these two real-time threshold values are due to the randomness of matched frames. 
Among a set of consecutive frames, selecting either one for loop closure may result in a slightly different visual transformation. 
For {real-time thresholds} of 70 and 100, the original approach is not able to detect any loop closure, because all the related nodes are transferred to LTM. 
But in our approach, we are able to retrieve back the transferred nodes to WM using Wi-Fi sensing. 
So we are able to make correct loop closures and experience a much smaller trajectory error, especially in A Hall and J Hall. 
These two datasets are very large and odometry measurements are subject to noise accumulation. 
Therefore, a correct loop closure benefits the localization accuracy to a high extent. 
Figure~\ref{fig:rtabmap_maps_2} presents the constructed maps of both approaches for J Hall in {real-time threshold}=70. % and A Hall 
For B Hall, the results show a lower error for vanilla RTAB-Map. 
We believe this stems from the small size of the dataset. 
Because in vanilla RTAB-Map, the trajectory error of B hall with no loop closure in {real-time threshold} of 70 and 100 is less than the case with correct detected loop closures with {real-time threshold}=$\infty$ when no frame is transferred to LTM. 
%This case approves our belief about the role of odometry in small environments.
In such an environment, the odometry may be able to provide a more accurate trajectory than visual estimations. 

Table~\ref{table:rtabmap_false_negative} shows the percentage of false negative loop closures of each approach for different {real-time threshold} values. 
For {real-time threshold}=200 and {real-time threshold}=$\infty$, the percentage of false negatives of both approaches is less than 20.
This is due to infrequent or no frame transfers to LTM.
Percentage values higher than zero in {real-time threshold}=$\infty$, where no frames are transferred to LTM, show that several loop closures are missed due to the random absence of some visual features in multiple frames.
As shown, all the possible correct loop closures are missing in the original approach for {real-time thresholds} of 70 and 100 due to unavailability of corresponding frames in WM. 
Two conditions lead to non-zero percentage values in our approach: 1- Random absence of visual features in some frames, 2- Availability of frames in WM; different pools of frames in WM generate different loop closure candidates.
% \zaki{Two conditions lead to non-zero percentage values in our approach: 1- Random absence of visual features, 2- Applying normalization and Bayes estimation for finding the best candidate for loop closure which is dependent on which frames are available in WM, and different subsets could lead to different calculations and candidate selection. %Time and position of pause points which could cause a small delay in frame retrieval from LTM.
%For avoiding the second condition, more frequent pause points could be solution and a trade off between operation runtime and accuracy should be adjusted.
But based on results in Table~\ref{table:rtabmap_rmse}, low percentage of false negatives doesn't affect the trajectory accuracy.
We don't see any specific reason for higher percentage values of A and J Hall datasets for {real-time threshold} of 100 except the above conditions.
Since there are no false positive loop closures detected in either approach in RTAB-Map, no results are shown.   
\begin{table*}
\caption{{\bf RTAB-Map:} Loop closure compute time (s) for different datasets with different {\it real-time thresholds} }
\begin{center}
\begin{tabular}{| c | r r | r r | r r | r r | } 
\hline 
\multicolumn{9}{|c|}{Real-time Threshold (ms)} \\ 
\hline 
 & \multicolumn{2}{|c|}{$\infty$} & \multicolumn{2}{|c|}{70} & \multicolumn{2}{|c|}{100} & \multicolumn{2}{|c|}{200} \\ 
\hline 
 Dataset & \multicolumn{1}{|c|}{Vanilla} & \multicolumn{1}{|c|}{WiFi} & \multicolumn{1}{|c|}{Vanilla} & \multicolumn{1}{|c|}{WiFi} & \multicolumn{1}{|c|}{Vanilla} & \multicolumn{1}{|c|}{WiFi} & \multicolumn{1}{|c|}{Vanilla} & \multicolumn{1}{|c|}{WiFi} \\ 
\hline 
 A Hall & 7.1138 & 2.3583 & 3.5781 & 5.0573 & 3.7496 & 5.0254 & 6.9523 & 3.9910 \\ 
 B Hall & 2.3625 & 0.6308 & 4.6772 & 1.8588 & 4.9314 & 1.8833 & 2.6194 & 1.6994 \\ 
 C Hall & 1.6906 & 0.6270 & 4.9086 & 1.6675 & 4.7170 & 1.6749 & 3.5015 & 1.3807 \\ 
 J Hall & 4.7277 & 0.8036 & 7.5797 & 2.6016 & 6.7597 & 2.6482 & 4.4986 & 2.8115 \\ 
\hline 
\end{tabular} 
\label{table:rtabmap_bounding}
\end{center}
%\vspace{-15pt}
\end{table*} 

In Table~\ref{table:rtabmap_bounding}, we show the computation time of loop closure detection for both approaches and all datasets. 
{Bounding loop closure} is supposed to save computation time in this process by restricting the number of comparisons. 
The results confirm that almost for all datasets and {\it real-time thresholds}, we are spending much less time for loop closure detection. 
The only different cases are for very long A dataset for {\it real-time thresholds} of 70 and 100. 
This is due to the size and shape of the environment. 
The A Hall dataset is very large and there are not many blocking walls along the trajectory which causes less RSSI attenuation between different places. 
This leads to less number of Wi-Fi clusters. 
As previously mentioned, there are only 8 Wi-Fi clusters created in this very large environment while the same number for smaller J and B datasets are 19 and 13. 
A low number of Wi-Fi clusters in a large environment lead to a very high number of keyframes in each cluster. 
In this situation, visual comparison to even one single {similar cluster} takes more time while the original approach is transferring many nodes to LTM due to the low {\it real-time threshold} and does a very low number of visual comparisons.

\begin{table}
\caption{{\bf RTAB-Map:} Compute overheadi (s) of {Wi-Fi clustering} and {cluster management} for different datasets and different {\it real-time thresholds}}
\begin{center}
\begin{tabular}{| c | r | r | r | r | } 
\hline 
\multicolumn{5}{|c|}{Real-time Threshold (ms)} \\ 
\hline 
 & \multicolumn{1}{|c|}{$\infty$} & \multicolumn{1}{|c|}{70} & \multicolumn{1}{|c|}{100} & \multicolumn{1}{|c|}{200} \\ 
\hline 
 {Dataset} & \multicolumn{1}{|c|}{WiFi} & \multicolumn{1}{|c|}{WiFi} & \multicolumn{1}{|c|}{WiFi} & \multicolumn{1}{|c|}{WiFi} \\ 
\hline 
 A Hall & 0.3208 & 0.3411 & 0.3432 & 0.3314 \\ 
 B Hall & 0.2129 & 0.2230 & 0.2203 & 0.2170 \\ 
 C Hall & 0.1701 & 0.1795 & 0.1804 & 0.1683 \\ 
 J Hall & 0.5633 & 0.6351 & 0.6261 & 0.6261 \\ 
\hline 
\end{tabular} 
\label{table:rtabmap_overhead}
\end{center}
%\vspace{-10pt}
\end{table}
Table~\ref{table:rtabmap_overhead} represents the computation overhead of our approach for all datasets caused by {\it Wi-Fi Clustering} and {\it Cluster Management} modules. 
As shown, these overheads are very small and could be ignored with respect to saved computation time shown in Table~\ref{table:rtabmap_bounding}. 
Based on the shown result, the amount of computation overhead is dependent on both the number of Wi-Fi clusters and the size of the dataset (number of frames). 
Higher number of clusters and higher number of frames lead to more comparisons and more computation overhead as expected. 
So although B dataset has a higher number of clusters than A dataset, it has a lower computation overhead due to less number of frames (748 compared to 1270).

\begin{table*}
\caption{{\bf ORB-SLAM:} Trajectory error (m) of different datasets and different {\it min-matches} values}
\begin{center}
\begin{tabular}{| c | r r | r r | r r | r r | r r |} 
\hline 
\multicolumn{11}{|c|}{{Min-Matches}} \\ 
\hline 
 & \multicolumn{2}{|c|}{10} & \multicolumn{2}{|c|}{15} & \multicolumn{2}{|c|}{20} & \multicolumn{2}{|c|}{50} & \multicolumn{2}{|c|}{100} \\ 
\hline 
 {Dataset} & \multicolumn{1}{|c|}{Vanilla} & \multicolumn{1}{|c|}{WiFi} & \multicolumn{1}{|c|}{Vanilla} & \multicolumn{1}{|c|}{WiFi} & \multicolumn{1}{|c|}{Vanilla} & \multicolumn{1}{|c|}{WiFi} & \multicolumn{1}{|c|}{Vanilla} & \multicolumn{1}{|c|}{WiFi} & \multicolumn{1}{|c|}{Vanilla} & \multicolumn{1}{|c|}{WiFi} \\ 
\hline 
 B Hall & 10.9616 & 0.2150 & 10.0935 & 0.2322 & 0.2513 & 0.2363 & 0.2111 & 0.2002 & 0.4189 & 0.4551 \\ 
 C Hall & 12.2219 & 0.1656 & 13.5045 & 0.1552 & 0.1177 & 0.0986 & 0.2129 & 0.2057 & 0.3366 & 0.3807 \\ 
 J Hall & 0.8638 & 0.8616 & 0.8234 & 0.7394 & 0.7407 & 0.6680 & 0.8564 & 0.7793 & 2.7230 & 2.8420 \\ 
 \hline 
 \end{tabular}
\label{table:orbslam_rmse}
\end{center}
%\vspace{-20pt}
\end{table*}
\subsection{ORB-SLAM performance}
%ORB-SLAM is one of the most recent SLAM algorithms. In this approach, they use a visual word dictionary in order to find loop closure candidates for the current frame. Although ORB-SLAM seems very accurate in map construction, but it yet could suffer from perceptual aliases. In order to show the applicability of our proposed method for ORB-SLAM, we try different {min-matches} values. This parameter specifies the minimum number of required inliers in RANSAC algorithm for accepting a visual transformation.

For ORB-SLAM, the tunable parameter is {\it min-matches} which is the number of inliers required in matching frames to accept a visual transformation. 
{\bf NOTE}: We were not able to get any result for A Hall in ORB-SLAM due to it having very low number of features.

Table~\ref{table:orbslam_rmse} shows the error of estimated trajectories for different {\it min-matches}. 
B Hall and C Hall experience false positive loop closures for {\it min-matches} of 10 and 15. 
So the estimated trajectories are inaccurate.
But we are able to avoid any false positive loop closure even with low values for {\it min-matches}. 
This shows that our proposed approach is applicable in symmetric environments having similar looking scenes. 
For higher values of {\it min-matches}, the trajectory error of both approaches is similar to each other.
In {\it min-matches}=100, the trajectory error is higher than in other settings. This is due to having a lower number of permissible visual transformations and more false negatives.
\begin{table*}
\caption{{\bf ORB-SLAM:} Loop closure compute time (s) for different datasets and different {\it min-matches} values. Our method reduces this time by 15\%-25\% on average. }
\begin{center}
\begin{tabular}{| c | r r | r r | r r | r r | r r |} 
\hline 
\multicolumn{11}{|c|}{{Min-Matches}} \\ 
\hline 
 & \multicolumn{2}{|c|}{10} & \multicolumn{2}{|c|}{15} & \multicolumn{2}{|c|}{20} & \multicolumn{2}{|c|}{50} & \multicolumn{2}{|c|}{100} \\ 
\hline 
 {Dataset} & \multicolumn{1}{|c|}{Vanilla} & \multicolumn{1}{|c|}{WiFi} & \multicolumn{1}{|c|}{Vanilla} & \multicolumn{1}{|c|}{WiFi} & \multicolumn{1}{|c|}{Vanilla} & \multicolumn{1}{|c|}{WiFi} & \multicolumn{1}{|c|}{Vanilla} & \multicolumn{1}{|c|}{WiFi} & \multicolumn{1}{|c|}{Vanilla} & \multicolumn{1}{|c|}{WiFi} \\ 
\hline 
 B Hall & 10.0983 & 7.0450 & 9.6854 & 7.2903 & 9.1030 & 6.5061 & 8.9114 & 7.1104 & 9.2383 & 7.6964 \\ 
 C Hall & 19.0397 & 16.4959 & 19.6789 & 15.9019 & 21.1756 & 18.8478 & 18.2757 & 15.7806 & 18.5043 & 15.5856 \\ 
 J Hall & 25.2880 & 19.6396 & 25.3739 & 20.0801 & 24.9643 & 19.9932 & 25.3299 & 20.3225 & 24.6863 & 19.7924 \\ 
\hline 
\end{tabular}  
\label{table:orbslam_bounding}
\end{center}
%\vspace{-20pt}
\end{table*}

In Table~\ref{table:orbslam_bounding}, we show the processing time of loop closure detection in the original approach and in our approach. 
Results show that for a similar level of accuracy in Table~\ref{table:orbslam_rmse}, we spend less time (15\%-25\% on average) for loop closure detection resulting in faster execution.
The difference in computation time grows larger as the dataset gets bigger(J Hall compared to B and C Hall). 
The reason is that the original approach always searches through all keyframes, but our approach is able to bound the searching process to keyframes of regions spatially close to current frame using Wi-Fi sensing. 
Further, the false positive loop closures increase the computation time by running global bundle adjustment wrongly. 
This shows that our Wi-Fi sensing is able to save time in the optimization process by avoiding false loop closures as well.  
\begin{table}[t]
\caption{{\bf ORB-SLAM:}Compute overhead (s) of {\it Wi-Fi Clustering} and {\it Cluster Management} modules for different datasets and {\it min-matches} values. It is about 3\%-8\% on average. }
\begin{center}
\begin{tabular}{| c | r | r | r | r | r |} 
\hline 
\multicolumn{6}{|c|}{{Min-Matches}} \\ 
\hline 
 & \multicolumn{1}{|c|}{10} & \multicolumn{1}{|c|}{15} & \multicolumn{1}{|c|}{20} & \multicolumn{1}{|c|}{50} & \multicolumn{1}{|c|}{100} \\ 
\hline 
{Dataset} & \multicolumn{1}{|c|}{WiFi} & \multicolumn{1}{|c|}{WiFi} & \multicolumn{1}{|c|}{WiFi} & \multicolumn{1}{|c|}{WiFi} & \multicolumn{1}{|c|}{WiFi} \\ 
\hline 
 B Hall & 0.4231 & 0.4638 & 0.4115 & 0.4543 & 0.5344 \\ 
 C Hall & 0.4929 & 0.3961 & 0.4498 & 0.6529 & 0.4667 \\ 
 J Hall & 1.7542 & 1.6599 & 1.5999 & 1.7060 & 1.7701 \\ 
\hline 
\end{tabular} 
\label{table:orbslam_overhead}
\end{center}
%\vspace{-10pt}
\end{table}

Table~\ref{table:orbslam_overhead} represents the computation time overhead caused by {\it Wi-Fi Clustering} and {\it Cluster Management} modules in our approach. 
As shown in Table~\ref{table:orbslam_bounding}, the computation time savings far outweigh these overhead values. 
These overhead values are dependent on the number of keyframes and Wi-Fi clusters. 
J dataset has a little more overhead due to more number of keyframes(about 2800) and 19 Wi-Fi clusters. 
B and C datasets are analogous to each other due to some randomness in their number of keyframes for different values of {\it min-matches}. 
For example for {\it min-matches}=50, the respective keyframes of B and C datasets are 1200 and 1500. 
So although C dataset has a lower number of Wi-Fi clusters, it has more computation overhead. 
\begin{table}[t]
\caption{{\bf ORB-SLAM:} Percentage false positive loop closures for different datasets and different {\it min-matches} values. Our method has zero false positives and is therefore not shown in the table. }
\begin{center}
\begin{tabular}{| c | r | r | r | r | r |} 
\hline 
\multicolumn{6}{|c|}{{Min-Matches}} \\ 
\hline 
 & \multicolumn{1}{|c|}{10} & \multicolumn{1}{|c|}{15} & \multicolumn{1}{|c|}{20} & \multicolumn{1}{|c|}{50} & \multicolumn{1}{|c|}{100} \\ 
\hline 
 {Dataset} & \multicolumn{1}{|c|}{Vanilla} & \multicolumn{1}{|c|}{Vanilla} & \multicolumn{1}{|c|}{Vanilla} & \multicolumn{1}{|c|}{Vanilla} & \multicolumn{1}{|c|}{Vanilla} \\ 
\hline 
 B Hall & 40.0000 & 33.3000 & 0.0000 & 0.0000 & 0.0000 \\ 
 C Hall & 50.0000 & 50.0000 & 0.0000 & 0.0000 & 0.0000 \\ 
 J Hall & 0.0000 & 0.0000 & 0.0000 & 0.0000 & 0.0000 \\  
\hline 
\end{tabular} 
\label{table:orbslam_false_positive}
\end{center}
%\vspace{-10pt}
\end{table}

In Table~\ref{table:orbslam_false_positive}, we show the percentage of false positive loop closures with respect to all detected ones in each parameter setting. 
As discussed earlier, B and C dataset suffer from some false positives for {\it min-matches} $\in (10, 15)$. %values of 10 and 15. 
This shows the vulnerability of the original approach to symmetric environments and demonstrates the ability of our version of ORB-SLAM to work well even when the {\it min-matches} value is low. 
Our approach using Wi-Fi does not incur any false positives for loop closure detection and is therefore not shown in the table. 
Also, there are no false negatives observed in any settings except for {\it min-match}=100 which is equal for both approaches.
%This could help compute time as it reduces the overhead of loop closure matching as shown in Table~\ref{table:orbslam_bounding} 
%i.e., for a similar level of accuracy, our algorithm could use a lower value for {\it min-matches} resulting in faster execution. 
%\vspace{-5pt}
\subsection{Comparison to Wi-Fi augmented FABMAP}
\begin{figure*}
	\begin{subfigure}[b]{.24\textwidth}
		\includegraphics[width=\textwidth]{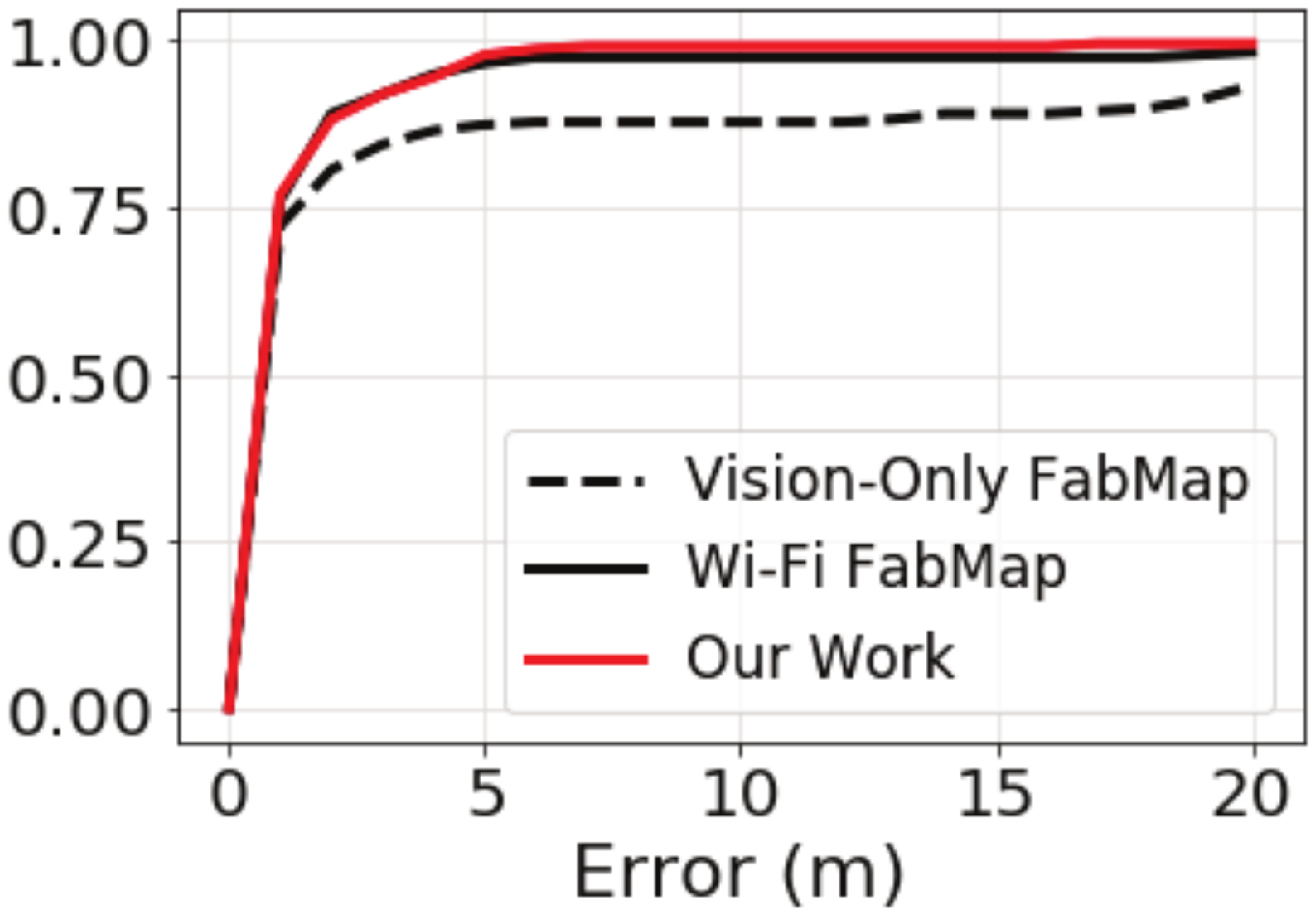}
		%\label{subfig:center}
		%\vspace{-6mm}
		\caption{C Hall}
	\end{subfigure}
	\begin{subfigure}[b]{0.24\textwidth}
		\includegraphics[width=\textwidth]{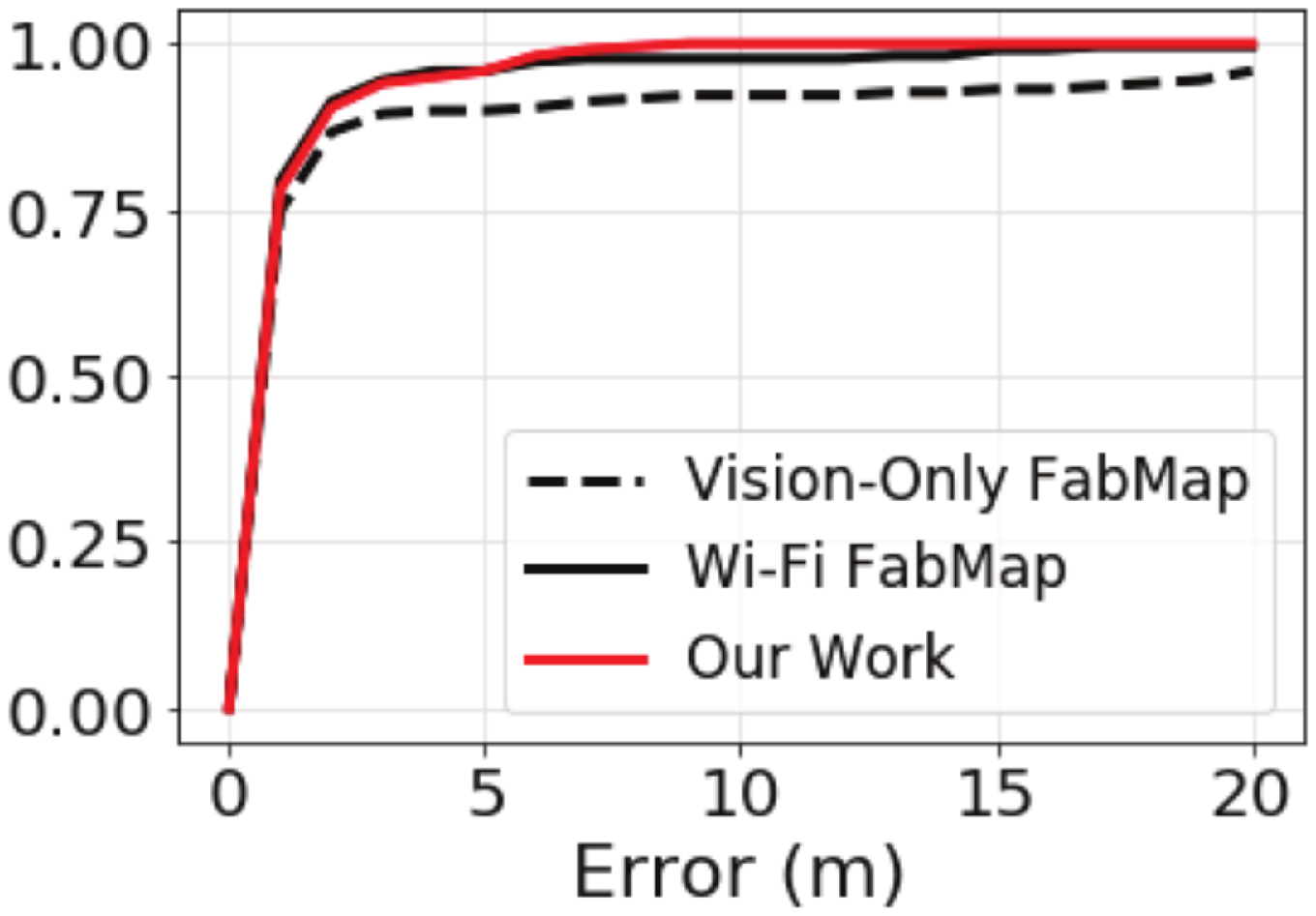}
		%\label{subfig:center}
		%\vspace{-6mm}
		\caption{B Hall}
	\end{subfigure}
	\begin{subfigure}[b]{0.24\textwidth}
		\includegraphics[width=\textwidth]{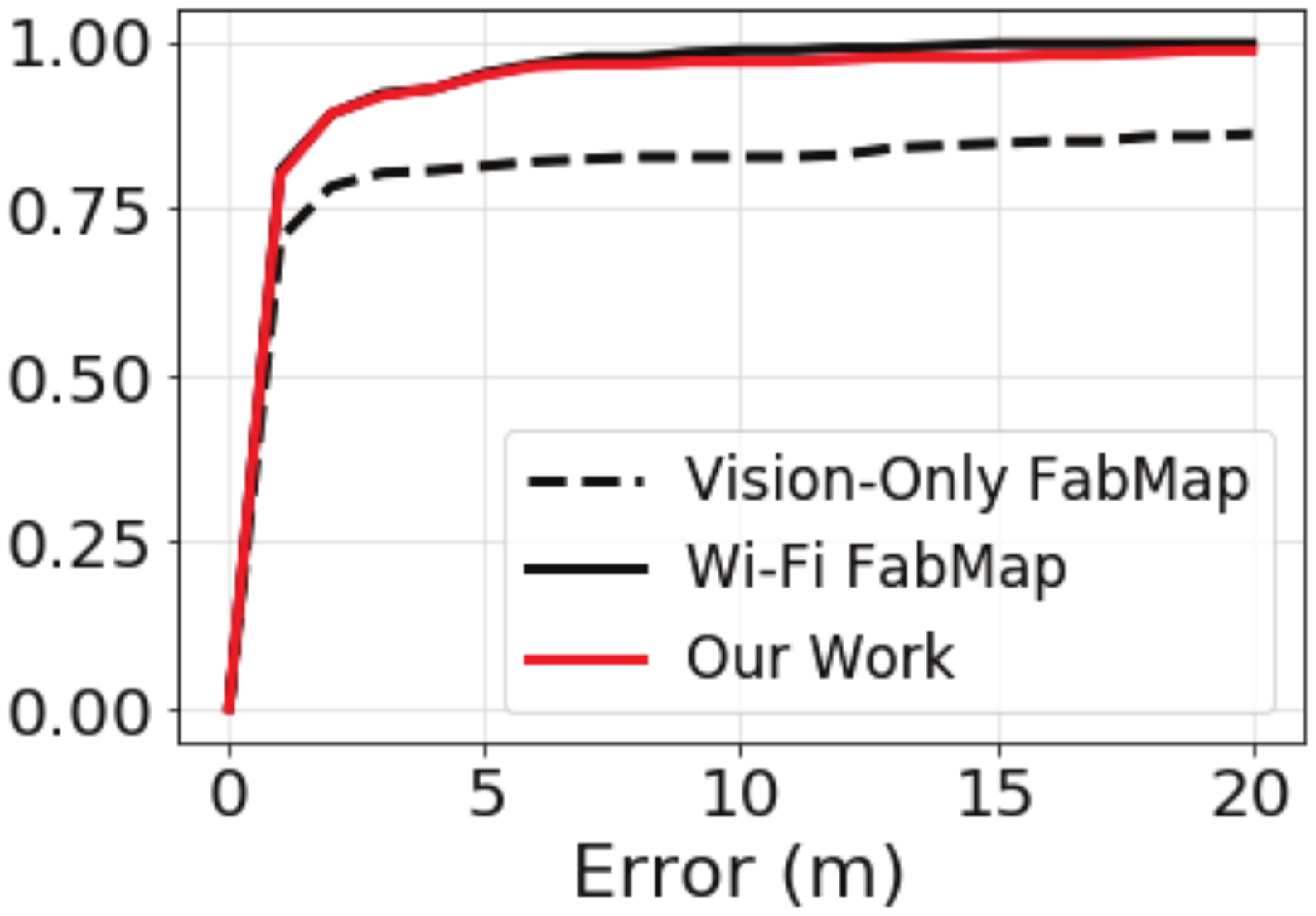}
		%\label{subfig:center}
		%\vspace{-6mm}
		\caption{J Hall}
	\end{subfigure}
	\begin{subfigure}[b]{0.24\textwidth}
		\includegraphics[width=\textwidth]{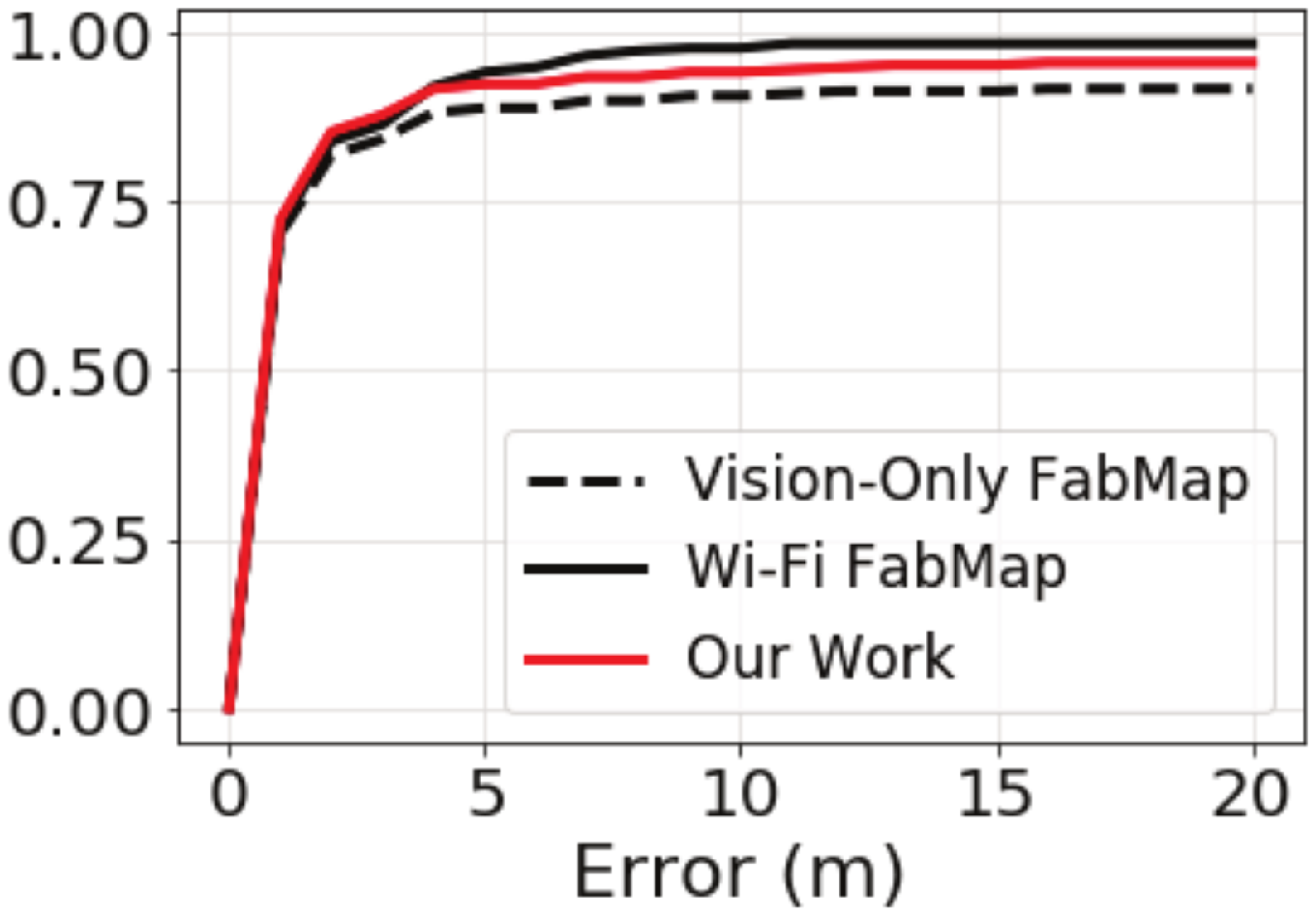}
		%\label{subfig:center}
		%\vspace{-6mm}
		\caption{A Hall}
	\end{subfigure}
%\vspace{-10pt}
\caption{CDF of Error in Wi-Fi augmented FABMAP compared to our approach}
\label{fabmap}
%\vspace{-10pt}
\end{figure*}
To situate our work wrt other visual SLAM algorithms that integrate Wi-Fi, we chose Wi-Fi FABMAP~\cite{visual_wifi_2}, the most recent work
that we came across that integrates wireless sensing with a specific visual SLAM algorithm. 

Wi-Fi augmented FABMAP~\cite{visual_wifi_2} is a topological localization algorithm which introduces a new approach for early fusion of visual and Wi-Fi information. It is executed in two phases: the mapping phase and the localization phase.
In the mapping phase, the robot is driven through the target area to collect Wi-Fi AP MAC addresses and images of the environment.
In this phase, each image is assumed to be from a different location and is associated with the spatially closest collected Wi-Fi vector, which is a binary vector indicating the presence of APs.
In the localization phase, the feature vector extracted from the query image is concatenated with the Wi-Fi vector collected at the location and fed into the FABMAP algorithm to find the best match among the images collected during the mapping phase.

{\bf Note:} (i) it is a two-phased method that requires war-driving. (ii) Wi-Fi FABMAP does not use signal strength values. Instead, it simply creates the Wi-Fi vector as a vector of binary values indicating the presence or absence of APs. (iii) It uses the Wi-Fi information only in the localization phase as opposed to integrating it into the SLAM process.

%\footnote{We spent some time on an available GitHub repo (\url{https://github.com/LRMPUT/WiFi-FAB-MAP}) but could not get it to work properly}
We faithfully re-implemented their algorithm by adapting the available open source code-base for FABMAP~\cite{glover2012openfabmap}. We collected data for input as per their paper by acquiring images every 2s and Wi-Fi data every 10s while the robot is in motion. The collected data is split into two sets: around 40\% for the mapping phase and 60\% for the localization phase. We also associated the Wi-Fi data with the images as per the procedure mentioned separately for the two phases.
The visual vocabulary was learned from 1500 images of corridor scenes from~\cite{yang_icra16,quattoni_cvpr09} and self-collected images in our university.

%Since topological SLAM, like FabMap, is different from metric SLAM,
FABMAP is a topological SLAM algorithm and not a metric SLAM algorithm such as RTAB-Map, and ORB-SLAM. Therefore, FABMAP and its derivatives would perform poorly in direct localization or mapping error comparison. For a fairer comparison, {\it we chose the exact same metrics} as used in~\cite{visual_wifi_2}. 
 %We chose to use the metric used in~\cite{visual_wifi_2} rather than the metrics used for RTABMap and ORB-SLAM comparisons. 
This is the CDF of distances between the estimated location and ground-truth averaged over the query images. To enable this comparison, we also provide localization results from our method over the query images rather than produce a trajectory.
In our approach, we use Wi-Fi signatures, which is a vector of RSSI values, rather than a binary vector used by Wi-Fi augmented FABMAP. We first find a representative Wi-Fi signature for each Wi-Fi cluster and the time stamp associated with it. %We assign this signature to all map images based on their time stamps. 
Then we associated every map image to a specific cluster based on their acquisition time stamps. For localization, the query image is also assigned a Wi-Fi signature. This is the signature recorded at the last pause prior to its acquisition. 
To get the best map image matching the query image, we first select clusters that their representative Wi-Fi signatures have high cosine similarity with the Wi-Fi signature of the query image named {\it similar clusters}. Then among the map images associated with similar clusters, we select the map image with maximum visual likelihood with the query image.%an image from the clusters based on the visual likelihood comparison with the query image.

%As mentioned previously, we use the metrics and illustrations similar to~\cite{visual_wifi_2} in this section. 
Figure~\ref{fabmap} shows the CDF of error between estimated localization and ground-truth of our Wi-Fi clustering method, Wi-Fi augmented FABMAP and visual FABMAP for all four datasets. We perform better for Halls B and C because:
\begin{itemize}
\item In some instances, the high similarity in visual features from physically different locations causes the algorithm to make the wrong matches even with the Wi-Fi data.
\item The constructed Wi-Fi Chow Liu tree could be inaccurate in capturing relations due to using raw Wi-Fi data.
The randomness associated in the detection of APs over time could cause the Chow Liu tree to have dependencies that don't necessarily hold.
Using our approach and aggregating the BSSIDs differing only in the last nibble could reduce the problem.
\end{itemize}
Our performance is similar to~\cite{visual_wifi_2} for J Hall. For Hall A, Wi-Fi augmented FABMAP performs better. The reason for this is that A Hall is a wide area with fewer features that doesn't include many blocking objects like walls along the trajectory. This causes less RSSI attenuation and therefore lesser number of Wi-Fi cluster. Since there are fewer clusters, there are more images per cluster and this increases the chances for perceptual aliasing. But we do note that even under these circumstances, more than 90\% of the query images are within 4 meters accuracy which is similar to Wi-Fi augmented FABMAP.

In summary, the demonstrated results show our performance improvement, in terms of localization/mapping accuracy and computational complexity, in visual SLAM algorithms.
% This improvement is due the employment of the unique features of Wi-Fi data in a well-suited and generic way in the original approaches.
The low dimensionality of Wi-Fi data and its immunity to perceptual aliasing are the key elements of the performance gain.
Further, the comparison with the state-of-the-art Wi-Fi FABMAP, shows a similar if not better performance.
This shows the generality of our approach unlike Wi-Fi augmented FABMAP which is specifically designed for visual FABMAP.

\vspace{-5pt}
\section{Discussion}\label{sec:discussion}
%In this work, we generalize the use of Wi-Fi sensing as a complementary modality to visual SLAM. We demonstrate this by incorporate Wi-Fi sensing into three distinct SLAM algorithms. Specifically, we observe that Wi-Fi sensing can provide coarse locality and incorporate a measure of this (Wi-Fi similarity) to improve SLAM. There are several factors that could affect the potential benefit from Wi-Fi sensing. Some are discussed below.
We incorporate Wi-Fi sensing into visual SLAM to combat two specific challenges of SLAM indoors --- perceptual aliasing and computational overhead. 
We discuss the relevance of our work to this community and the implications of some of our choices here. 

{\bf Relevance to Sensor Systems:} Our work is useful for robots as well as mobile devices. 
With the advent of RGB-D cameras for mobile devices such as Intel RealSense and augmented reality/mixed reality devices such as the MS Hololens and MagicLeap One, we expect an increase in spatial applications that will use visual SLAM. Therefore, we believe that this topic is of relevance to both robotics and the sensor systems communities. 

{\bf Environmental Dynamics:} Wi-Fi signal strength can vary with environmental dynamics such as the number of people in the area, the number of devices connected etc.
However, our empirical observation from many repeated data collection trials is that this does not have a significant impact on our similarity measure.
%\zaki{We decided to use RSSI instead of CSI, because CSI is much more sensitive to any small dynamics in the environment to the extent of being used for gesture detection~\cite{}.}

{\bf Wi-Fi Similarity Tuning:} This parameter is analogous to the {\it min-matches} parameter in visual slam algorithms. 
We tried many values to find the optimal initialization which seems dependent on the size and the degree of dynamics of the environment. 
Spaces including more dynamics like A Hall dataset with many people in motion require lower values in order to compensate for fluctuations.
In general, very high values would increase the number of false negative loop closures and very low values would make it inapplicable for getting rid of perceptual aliases.

{\bf Number of Access Points:} Depending on the placement and number of visible access points, the effectiveness of Wi-Fi sensing might vary.
The performance gain would increase with higher number of APs especially if they are scattered and not co-linear. Based on our datasets, which reasonably represent modern urban settings, our approach works well with as low as 40 APs scattered around a square shaped environment.
%

%Depending on the deployment, the number of visible access points, and corresponding ability to sense Wi-Fi might vary. We have collected data from four university buildings that are reasonably representative of modern urban settings. Our approach works where there are at least three APs which is a reasonable assumption these days. 
%
 
% In summary, Wi-Fi sensing provides coarse locality to mapping. This is potentially useful to improve accuracy (in RGBD SLAM, for example) where the algorithm cannot determine this, and/or running time by comparing to only the relevant frames (in RTAB-Map in some cases). It is also a great tool to determine long-term loop closure. Given the low overhead of computation of Wi-Fi similarity in comparison to visual similarity, we believe that that Wi-Fi sensing could be a useful auxiliary sensing modality for indoor mapping. 

%\vspace{-5pt}
\section{Conclusion}
\label{sec:conc}
%\vspace{\vertspcposthead}
%In this paper, we proposed three variants to the RGBD SLAM algorithm, each of which modified the algorithm to use RSSI information from commodity-off-the-shelf Wi-Fi cards to disambiguate different locations in an indoor office environment that is repetitive or feature-less. We demostrated the quality of our method by comparing the results produced against the results produced by existing RGBD SLAM algorithm. Results from our experiments show that our variants produce trajectories that are up to 18 times more accurate than RGBD SLAM and have the added benefit of having the computation time reduced by up to 70\%.
%\vspace{-15pt}
%\vspace{-35pt}
%\vspace{-35pt}
In this work, we proposed a general approach to incorporate Wi-Fi sensing into visual SLAM algorithms. To demonstrate, 
we augment three recent SLAM algorithms to show improved mapping/localization accuracy as well as speed-up in operation. 
We demonstrated this functionality on data collected from four university buildings, which are representative spaces of modern urban environments.

We also compared our proposed approach to recently proposed Wi-Fi augmented FABMAP and showed a comparable if not better performance.
This comparison also confirms the generality of our approach unlike Wi-Fi augmented FABMAP which is only designed for visual FABMAP.

In the future, we hope to demonstrate the utility of Wi-Fi sensing for sustained long-term use in an urban space. 
While Wi-Fi signal strength used for this work is a useful measure, novel wireless sensing technologies such as 60 GHz sensing can be used to further improve the overall accuracy as well as the computational complexity of SLAM algorithms in the future as well.
%more detailed properties such as Channel State Information (CSI) can be obtained using modern Wi-Fi cards. These would greatly enhance Wi-Fi sensing ability and its corresponding use for multi-robot applications. We will explore using CSI for SLAM in the future.

%\begin{abstract}
%Insert your abstract here. Include keywords, PACS and mathematical
%subject classification numbers as needed.
%\end{abstract}

%\begin{acknowledgements}
%\end{acknowledgements}


%\begin{thebibliography}{}
\bibliographystyle{apalike}
\bibliography{wifislam}
%\newpage
%\input{bio}

% etc
%\end{thebibliography}

\end{document}